\documentclass[letterpaper,twocolumn,10pt]{article}
\usepackage{usenix}

\usepackage[all=normal,title=normal,floats=tight,paragraphs=tight]{savetrees}
\usepackage[compact,small]{titlesec}

\usepackage{algorithm}
\usepackage{algpseudocode}
\usepackage{amsmath}
\usepackage{amssymb}
\usepackage{amsfonts}
\usepackage{array} 
\usepackage{booktabs}
\usepackage{contour}
\usepackage{cuted}
\usepackage{enumitem} 
\usepackage{graphicx}
\usepackage{bbm}
\usepackage{multirow} 
\usepackage{pifont} 
\usepackage{soul}
\usepackage{float}
\usepackage{dsfont}
\usepackage{microtype}
\usepackage{subcaption}
\usepackage{tikz}
\usepackage{pifont}
\usetikzlibrary{shapes.geometric}
\usetikzlibrary{math}
\usepackage{titlesec}
\usepackage{subcaption}
\usepackage{xcolor}
\usepackage{afterpage}
\usepackage{filecontents}
\usepackage{multicol}
\usepackage{tabularx}

\titleformat{\paragraph}[runin]{\normalfont\normalsize\bfseries}{\theparagraph}{0em}{}

\newif\ifdraft 
\drafttrue      

\newbool{IsPrintComment}
\booltrue{IsPrintComment}
\newcommand{\da}[1]
{
    \ifbool{IsPrintComment}
    {%
        {\bf \color{orange} DA: #1}
    }{}
}
\newcommand{\algo}{Program-Level Attained Service}
\newcommand{\name}{\text{Autellix}}

\begin{document}
\date{}
\title{\name: An Efficient Serving Engine for LLM Agents as General Programs}
\author{
\rm{Michael Luo$^{\text{1,2}}$ \enskip
    Xiaoxiang Shi$^{\text{3}, \dagger}$ \enskip
    Colin Cai$^{\text{1}, \dagger}$ \enskip
    Tianjun Zhang$^{\text{1}}$ \enskip
    Justin Wong$^{\text{1}}$ \enskip}
\\
\rm{Yichuan Wang$^{\text{1}}$ \enskip
    Chi Wang$^{\text{2}}$ \enskip
    Yanping Huang$^{\text{2}}$ \enskip
    Zhifeng Chen$^{\text{2}}$ \enskip
    Joseph E. Gonzalez$^{\text{1}}$ \enskip
    Ion Stoica$^{\text{1}}$ \enskip}\\
\\
{$^{\text{1}}$UC Berkeley\enskip $^{\text{2}}$Google DeepMind\enskip$^{\text{3}}$Shanghai Jiao Tong University}
}

\maketitle

\def\thefootnote{†}\footnotetext{Equal and significant contribution.}\def\thefootnote{\arabic{footnote}}

\vspace{-10mm}
\pagestyle{empty}
\begin{abstract}

Large language model (LLM) applications are evolving beyond simple chatbots into dynamic, general-purpose agentic programs, which scale LLM calls and output tokens to help AI agents reason, explore, and solve complex tasks. However, existing LLM serving systems ignore dependencies between programs and calls, missing significant opportunities for optimization.
Our analysis reveals that programs submitted to LLM serving engines experience long cumulative wait times, primarily due to head-of-line blocking at both the individual LLM request and the program.

To address this, we introduce \text{\name}, an LLM serving system that treats programs as first-class citizens to minimize their end-to-end latencies. \text{\name} intercepts LLM calls submitted by programs, enriching schedulers with program-level context. We propose two scheduling algorithms—for single-threaded and distributed programs—that preempt and prioritize LLM calls based on their programs' previously completed calls. Our evaluation demonstrates that across diverse LLMs and agentic workloads, \text{\name} improves throughput of programs by 4-15× at the same latency compared to state-of-the-art systems, such as vLLM.

\end{abstract}

\section{Introduction}
\label{sec:intro}


Large language models (LLMs) as autonomous agents enhance their problem solving capabilities by scaling their inference computation—that is, increasing the number of output tokens or LLM calls~\cite{Evans1984-EVAHAA, Kahneman:2011fj, 01670ebe-b3af-3829-a929-8f19529d1afb, snell2024scalingllmtesttimecompute, chen2024llmcallsneedscaling, brown2024largelanguagemonkeysscaling}. With more calls and tokens, LLMs endow agents with improved reasoning~\cite{wei2023chainofthoughtpromptingelicitsreasoning, yao2023treethoughtsdeliberateproblem, yao2023reactsynergizingreasoningacting,deepseekai2025deepseekr1incentivizingreasoningcapability}, planning and search capabilities~\cite{zhou2024languageagenttreesearch, putta2024agentqadvancedreasoning}, self-reflection from prior experiences ~\cite{shinn2023reflexionlanguageagentsverbal, yu2024exactteachingaiagents, kumar2024traininglanguagemodelsselfcorrect}, and collaboration between multiple agents~\cite{du2023improvingfactualityreasoninglanguage, wu2023autogenenablingnextgenllm, zhuge2024languageagentsoptimizablegraphs}. These techniques enable agents to effectively navigate external environments via tools~\cite{schick2023toolformerlanguagemodelsteach, patil2023gorillalargelanguagemodel, yao2023reactsynergizingreasoningacting} and solve complex tasks, such as autonomously browsing the web~\cite{yao2023webshopscalablerealworldweb, zhou2024webarenarealisticwebenvironment, gur2024realworldwebagentplanninglong}, resolving GitHub issues~\cite{jimenez2024swebench, yang2024sweagentagentcomputerinterfacesenable, wang2024openhandsopenplatformai}, and proving difficult math problems~\cite{deepmind_imo_silver_2024, kumarappan2024leanagentlifelonglearningformal}. 

\begin{figure}[t]
\centering
\begin{subfigure}[b]{0.49\columnwidth}
    \includegraphics[width=\linewidth]{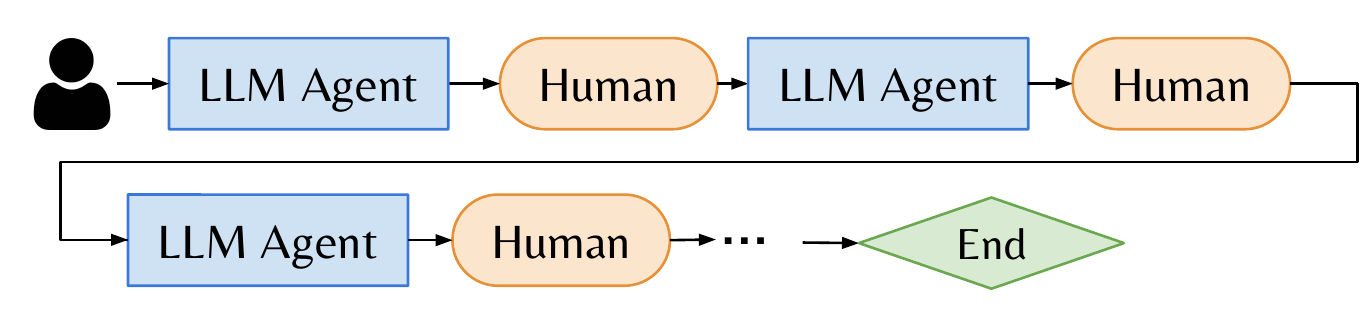}
    \caption{Chatbot}
    \label{fig:chatbot_agent}
\end{subfigure}
\hfill
\begin{subfigure}[b]{0.49\columnwidth}
    \includegraphics[width=\linewidth]{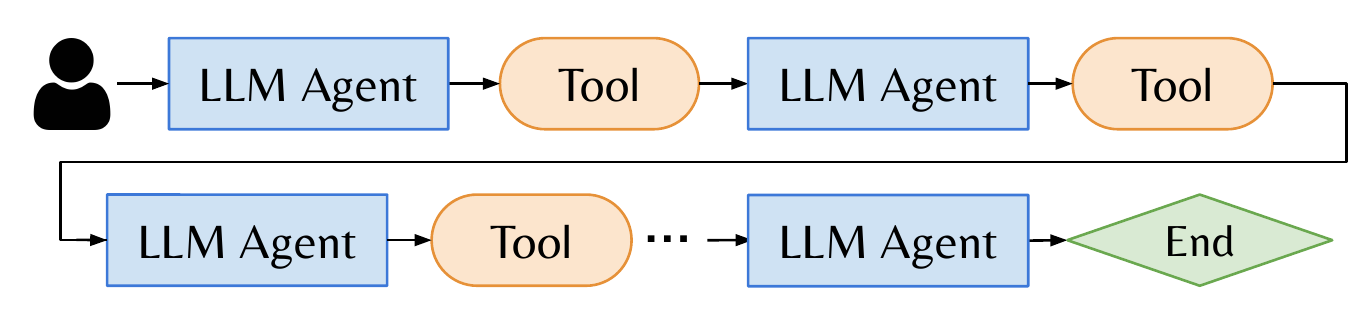}
    \caption{ReAct Agent}
    \label{fig:react_agent}
\end{subfigure}
\hfill
\begin{subfigure}[b]{0.44\columnwidth}
    \includegraphics[width=\linewidth]{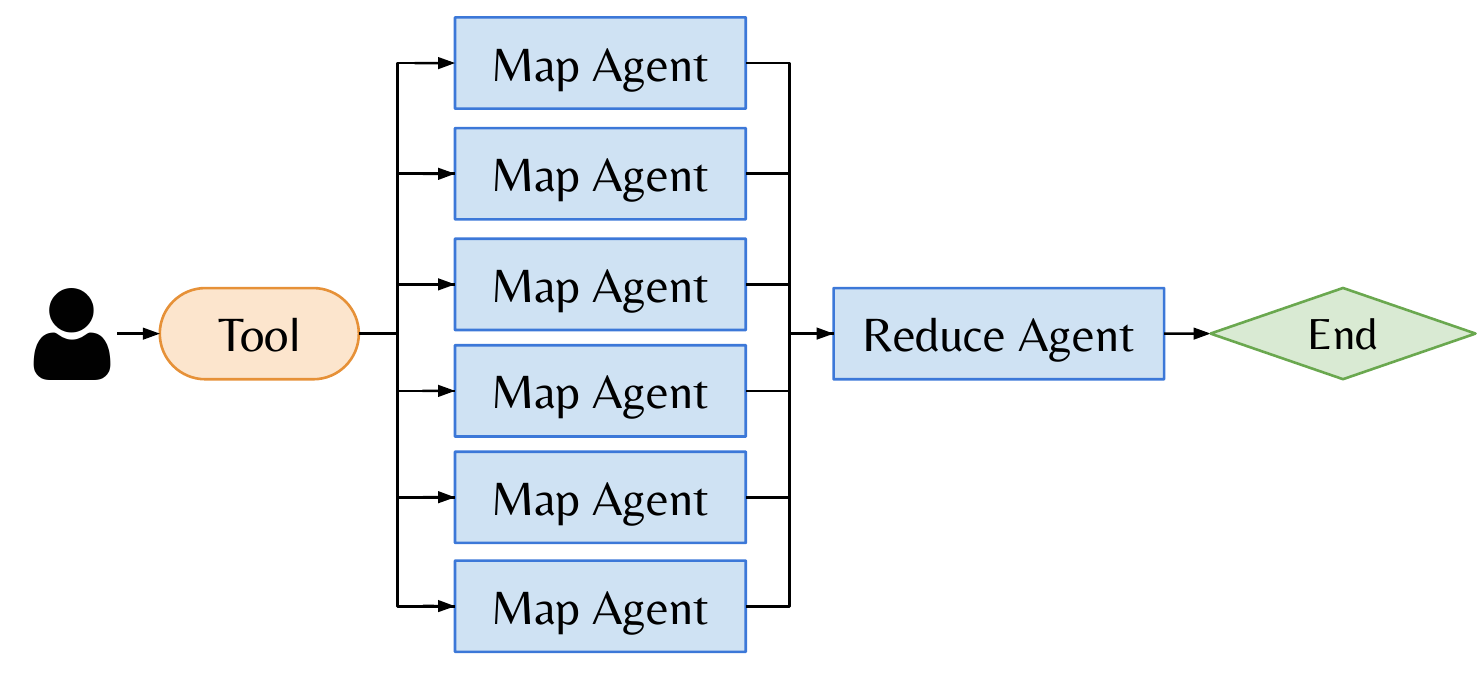}
    \caption{Map-Reduce}
    \label{fig:map_reduce_agent}
\end{subfigure}
\hfill
\begin{subfigure}[b]{0.53\columnwidth}
    \includegraphics[width=\linewidth]{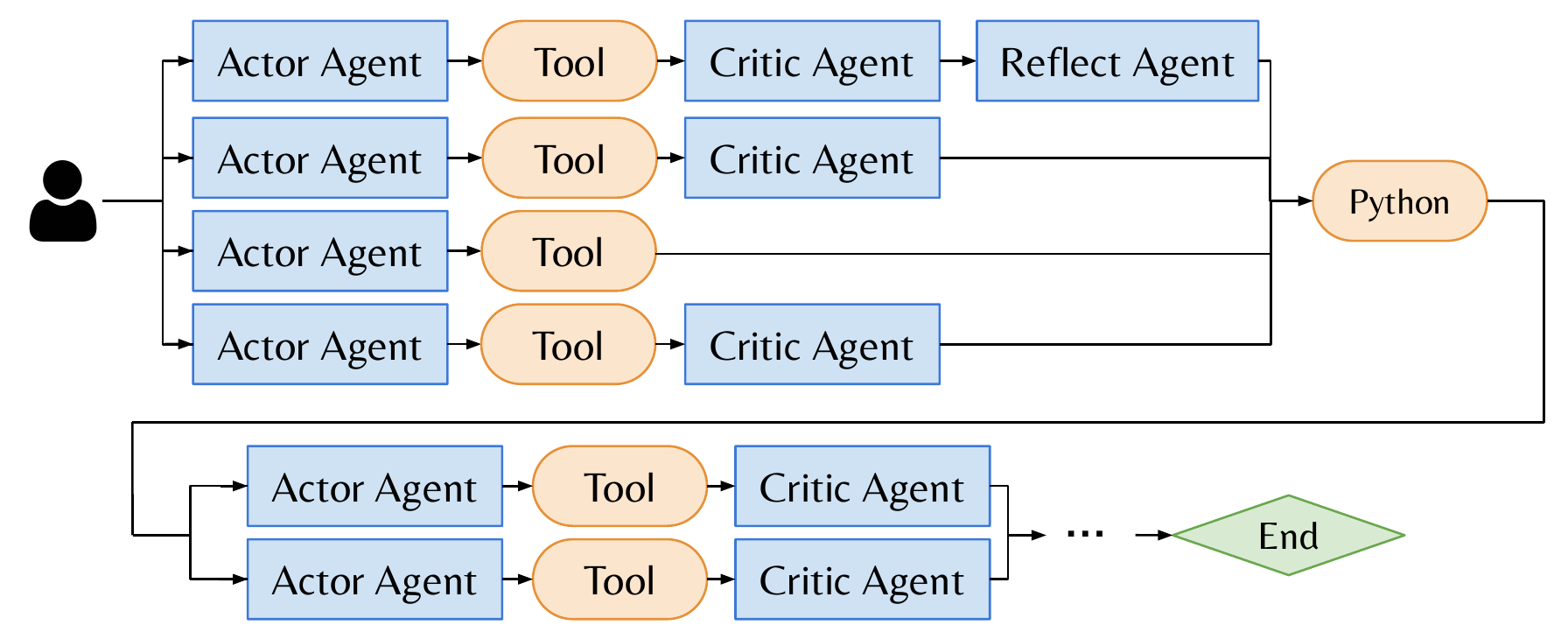}
    \caption{Monte Carlo Tree Search}
    \label{fig:mcts_agent}
\end{subfigure}
\caption{\small\textbf{Execution workflows for \textit{Agentic Programs}.} Agentic programs are highly dynamic execution workflows that follow a directed acyclic graph (DAG). It consists of \textcolor{blue}{LLM calls} from one or more LLM agents and \textcolor{orange}{external interrupts} (i.e. tool calls, humans).}
\label{fig:agentic_program}
\vspace{-5mm}
\end{figure}


\begin{figure*}[!htb]
\begin{subfigure}{0.5\textwidth}
    \centering
    \begin{tabular}{|c|c|c|}
        \hline
        \textbf{Program} & \textbf{\# LLM Calls}  & \textbf{Decode Steps per LLM Call} \\ 
        \hline
        A & 4 & \{4,3,1,1\} \\
        B & 3 & \{3,3,4\} \\ 
        C & 2 & \{1,2\} \\
        D & 1 & \{4\} \\
        \hline
    \end{tabular}
    \vspace{3mm}
    \caption{Programs, arriving at t=0}
    \label{fig:program_runtimes}
\end{subfigure}\hfill
\begin{subfigure}{.48\textwidth}
    \centering
    \includegraphics[width=\textwidth]{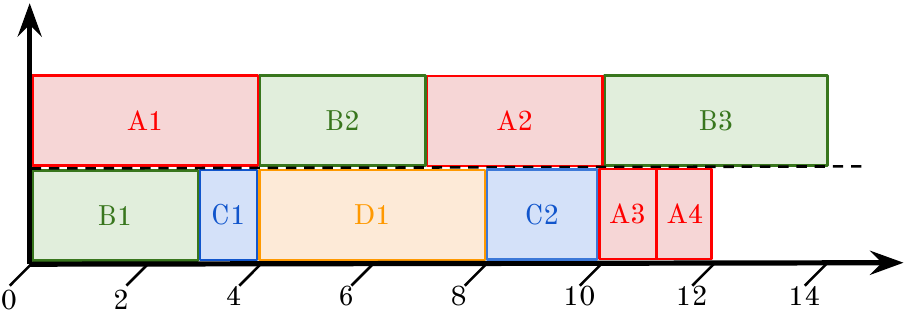}
    \caption{First-Come First-Served (FCFS)}
    \label{fig:fcfs_example}
\end{subfigure}\hfill
\vspace{-3mm}
\begin{subfigure}{.48\textwidth}
    \centering
    \includegraphics[width=\textwidth]{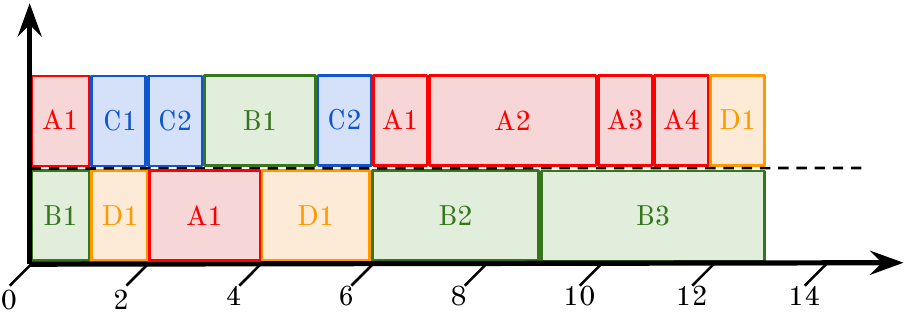}
    \caption{Multilevel Feedback Queue (MLFQ)}
    \label{fig:mlfq_example}
\end{subfigure}\hfill
\begin{subfigure}{.48\textwidth}
    \centering
    \includegraphics[width=\textwidth]{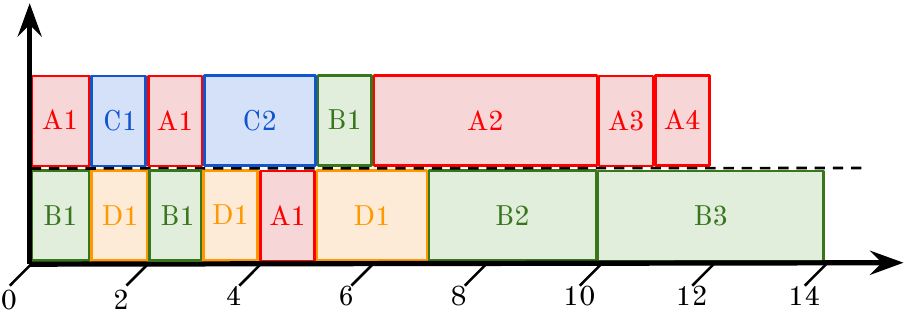}
    \caption{\text{\algo} (PLAS)}
    \label{fig:agentix_example}
\end{subfigure}

\vspace{-2.5mm}
\caption{\small \textbf{Gantt chart of LLM call execution on an LLM serving engine with a max batch size (BS) of 2 (Y-axis) over decoding steps (X-axis).}  
(a) Four programs vary in the number of LLM calls and decode steps per call. Long programs (A, B) and short programs (C, D) are shown.  
(b) First-Come First-Served (FCFS) incurs \textit{head-of-line blocking} as long LLM calls delay short LLM calls, resulting in a waiting time of \textbf{18 units}.  
(c) Multilevel Feedback Queue (MLFQ) reduces blocking with preemption but still incurs \textit{program-level} blocking. Programs A and B's new LLM calls are placed in the highest priority queue, delaying Program D, incurring \textbf{18 units} of waiting time.  
(d) \text{\algo} (PLAS) leverages program-level statistics, delaying subsequent calls in A and B to prioritize programs C and D, reducing waiting time to \textbf{12 units}.}
\vspace{-5mm}
\label{fig:tutorial_example}
\end{figure*}

The rise of inference-time techniques and agentic applications signifies a shift from static, specialized LLM applications~\cite{langchain, lin2024parrotefficientservingllmbased} to highly dynamic, general \textit{agentic programs}~\cite{zheng2024sglangefficientexecutionstructured,wu2023autogenenablingnextgenllm, langgraph}. More precisely, an agentic program is a dynamic execution workflow, represented by a directed acyclic graph (DAG), that consists of LLM calls from one or more agents, and external interrupts, which include tool calls (i.e. external API calls), generic code execution, or human inputs (\S\ref{sec:background}). We assume that the LLM invocation pattern of programs emerges only at runtime, making it difficult to fully know or predict the entire graph in advance.

Figure~\ref{fig:agentic_program} illustrates the highly dynamic nature of agentic programs with single and multi-threaded examples. Single-threaded programs vary in two dimensions: 1) the length of the program, which depends on the user prompt, and 2) the sequence of LLM calls and interrupts, determined by a program's control flow. For instance, both Chatbot and ReAct (Reasoning and Acting)~\cite{yao2023reactsynergizingreasoningacting} agents cycle between LLM calls and interrupts (human or tool call) and terminate based on a human or LLM's decision. (Fig.~\ref{fig:chatbot_agent},~\ref{fig:react_agent})~\cite{yao2023reactsynergizingreasoningacting}. Multi-threaded programs generally form DAGs. Both Map-Reduce, a classic multi-threaded program, and Monte Carlo Tree Search (MCTS) vary in the number of threads that fork and merge over time, where each thread may contain different sequences of LLM calls and interrupts (Fig.~\ref{fig:map_reduce_agent},~\ref{fig:mcts_agent}). In particular, MCTS is a widely used technique for search and planning for reasoning and web-based agents~\cite{zhou2024languageagenttreesearch, putta2024agentqadvancedreasoning, sequoia2024generative, muennighoff2025s1simpletesttimescaling}.

Existing LLM serving engines, like vLLM~\cite{vllm}, focus on optimizing individual LLM calls or static LLM applications~\cite{lin2024parrotefficientservingllmbased} by improving key-value (KV) cache efficiency~\cite{vllm, zheng2024sglangefficientexecutionstructured}, accelerating CUDA kernels~\cite{nanoflow,fastserve}, and better scheduling algorithms for LLM requests~\cite{fastserve, agrawal2024tamingthroughputlatencytradeoffllm}. However, these optimizations fail to account for the program-level context, such as the dependencies between LLM calls in the same program or program-level statistics, like total execution time. As a result, these systems often suffer from suboptimal end-to-end performance for complex programs—in particular, programs' end-to-end latencies (\S\ref{sec:motivation}).

Figure~\ref{fig:tutorial_example} illustrates a burst of two long programs (A, B) and two short programs (C, D) submitted to an LLM serving engine with a max batch size of 2 at t=0. Each program has one or more LLM calls with varying decoding lengths in Fig.~\ref{fig:program_runtimes}. 
Under a program-agnostic First-Come-First-Served (FCFS) policy, 
the default policy 
for vLLM~\cite{vllm}, long LLM calls block other calls from running, resulting in 
\textit{call-level} head-of-line (HoL) blocking, as shown in Fig.~\ref{fig:fcfs_example}. Program A and B's initial, long LLM calls execute first, delaying program C and D's execution until t=3,4. Repeated cases of HoL blocking result in a total waiting time of \textbf{18} units. To address this, preemptive scheduling, such as Multi-Level Feedback Queue (MLFQ)~\cite{fastserve}, reduces HoL blocking by preempting long LLM calls to let short calls execute. However, without program-level context, newer programs are repeatedly delayed by subsequent calls from older programs, incurring \textit{program-level} HoL blocking. In Fig.~\ref{fig:mlfq_example}, MLFQ successfully preempts program A and B's long calls to start executing C and D. However, MLFQ repeatedly prioritizes A and B's subsequent calls from t=6-12, which delays program D's execution. Consequently, MLFQ incurs the same wait time of \textbf{18} units as FCFS.

We present \text{\name}, an LLM inference system designed to run programs, not individual LLM calls. Inspired by OS schedulers for processes, our key idea is to prioritize LLM calls by the total execution time of their program's previously completed calls; LLM calls from long programs, which are unlikely to complete soon, are deprioritized, allowing shorter programs to complete first. In Fig.~\ref{fig:agentix_example}, short programs C and D are no longer blocked by subsequent LLM calls from long programs A and B, effectively eliminating HoL blocking and reducing the total wait time to \textbf{12} units. 

\text{\name} introduces a novel framework that leverages global, program-level statistics, such as program's cumulative execution time on an engine, to minimize waiting times and improve engine throughput. We propose two non-clairvoyant scheduling algorithms that assume no prior workload knowledge of programs: \textit{PLAS} (Program-Level Attained Service) for single-threaded programs and \textit{ATLAS} (Adaptive Thread-Level Attained Service) for multi-threaded programs represented as general, dynamic DAGs. \textit{PLAS} prioritizes LLM calls based on the current cumulative service, or execution times, of their source program.
Generalizing \textit{PLAS}, \textit{ATLAS} prioritizes LLM calls based on the maximum cumulative service time across all threads in the same program, which sorts calls based on their program's critical path~\cite{cpm}. Beyond reduced wait times, \textit{ATLAS} decreases program's makespans by prioritizing critical LLM calls that would otherwise block programs' progress (\S\ref{sec:solution}).

Programs comprised of tens to hundreds of LLM calls impose significant demands to the serving systems with a single LLM engine capable of handling only 0.2 programs per second for MCTS (\S\ref{sec:evaluation}). Hence, \text{\name} also routes programs' LLM calls across multiple engines. For agentic workloads, our key observation is that LLM calls within a program often share common prefixes and cumulative conversation states, while calls across programs typically share only the system prompt~\cite{preble}. To avoid recomputing the programs' KV-cache, \text{\name} respects a program's data locality by routing long calls to their programs' engines, while load-balancing shorter calls to other engines, where system prompts make up most of the input for shorter calls.

We implement a system prototype of \text{\name} as a layer on top of LLM serving engines, such as vLLM~\cite{vllm}, and expose a stateful API that allows users to establish persistent sessions with \text{\name}, unlike traditional stateless APIs~\cite{openai_chat_api}. We evaluate \text{\name} across different LLMs and four representative agentic workloads (\S\ref{sec:evaluation}). Our results show that \text{\name} improves throughput by 4-15x compared to state-of-the-art inference systems like vLLM~\cite{vllm}. Across engines, \text{\name} improves throughput by up to 1.5x over standard load-balancers.

\noindent In summary, the primary contributions of this paper are:
\begin{itemize}[itemsep=0pt, parsep=0pt, topsep=0pt, partopsep=0pt, leftmargin=*]
  \item This work is the first to formalize agentic programs as dynamic, non-deterministic DAGs of LLM calls and interrupts. (\S\ref{sec:background}) 
  \item \text{\name} utilizes program-level statistics to better inform its scheduler. \text{\name}'s non-clairvoyant scheduler requires only the cumulative service times of LLM calls within the same program. (\S\ref{sec:solution})
  \item \text{\name} leverages a simple load-balancing policy across multiple engines to balance data locality and KV-cache recomputation. (\S\ref{sec:solution})
  \item Our system is easily deployable, seamlessly integrates a stateful API with existing programming and agent frameworks, and  demonstrates significant throughput gains (\S\ref{sec:evaluation}).
\end{itemize}


\section{Background \& Related Work}
\label{sec:background}

To detail relevant context for \text{\name}, we provide a brief overview of the emergent AI agent infrastructure and its applications, split between the LLM serving layer (\S~\ref{sec:llm_serving}) and higher-level agentic layer  (\S~\ref{sec:llm_agents}), as depicted in Figure~\ref{fig:agent_infra}.

\subsection{LLM Serving Layer}
\label{sec:llm_serving}
\noindent \textbf{LLM Inference Process.} Large language models (LLMs), which drive chatbots and AI applications, predominantly utilize the Transformer architecture~\cite{transformer}, including decoder-only models such as GPT, Claude, and LLaMA~\cite{gpt,llama,llama2,mistral}. For each request, LLM inference operates in two stages: the \textit{prefill} phase, which converts the input prompt into intermediate token states, and the \textit{decoding} phase, where new tokens are generated auto-regressively, one at a time, based on prior token sequences. To reduce computation, LLM serving systems leverage \textit{KV-cache}, which stores intermediate token states to accelerate token generation~\cite{orca, vllm}.

\begin{figure}[t]
    \centering
    \includegraphics[width=\linewidth]{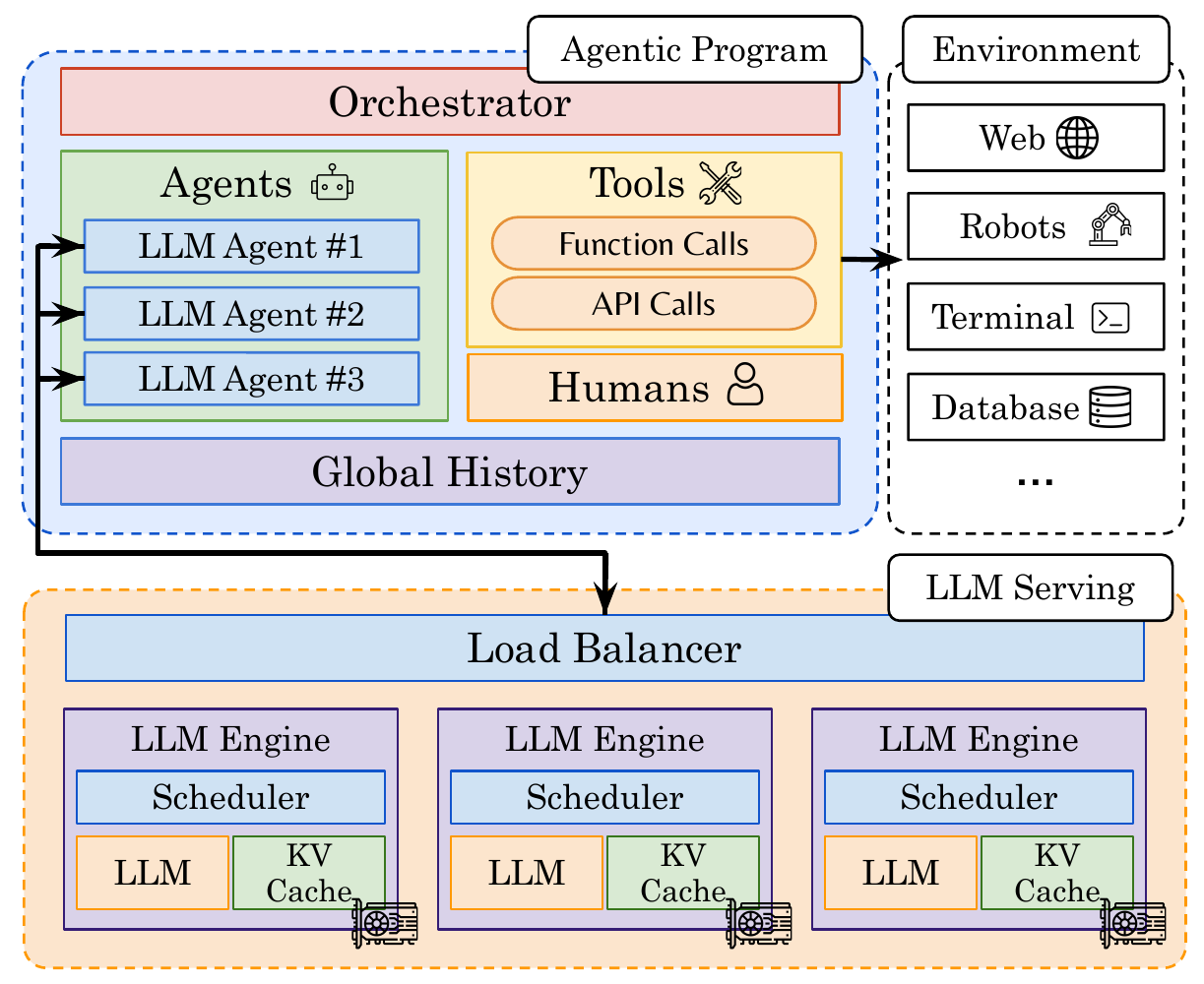}
    \caption{\small\textbf{AI Agent Infrastructure.} Top: Developers and users build and execute agentic programs that orchestrate execution and persist global, cumulative history across agents, tools, and humans. Bottom: LLM serving systems process agents' LLM calls and route calls across one or more LLM engines.}
    \label{fig:agent_infra}
     \vspace{-6mm}
 \end{figure}

\vspace{2mm}
\noindent \textbf{LLM Serving.} LLM serving systems manage both the routing of LLM calls across engines and the execution of LLM calls within each engine (Fig.~\ref{fig:agent_infra}). Within an engine, recent innovations in LLM serving mirror concepts rooted in traditional operating systems (OS), such as memory management, kernel optimization, and scheduling~\cite{mei2024aiosllmagentoperating, sun2024llumnixdynamicschedulinglarge}. Existing solutions, such as vLLM, integrate virtual memory and paging techniques to reduce KV-cache fragmentation~\cite{vllm}, introduce shared memory to cache prefixes across LLM requests~\cite{lin2024parrotefficientservingllmbased,zheng2024sglangefficientexecutionstructured}, and manage cache hierarchies between GPU, CPU, and disk~\cite{nanoflow, flexgen, slora}. Other techniques improve GPU kernel implementations to accelerate self-attention~\cite{flashattention}, pipeline different operators~\cite{nanoflow}, and implement better tensor or pipeline parallelism~\cite{fastserve, alpaserve}. Finally, LLM engines can leverage better scheduling, such as binpacking prefills and decodes together~\cite{agrawal2024tamingthroughputlatencytradeoffllm} and preempting LLM requests~\cite{fastserve}, to improve response times. Across multiple LLM engines, serving systems employ load-balancing techniques like live migration~\cite{llumnix}, disaggregate prefills and decodes~\cite{mooncake}, construct prefix trees~\cite{preble}, and migrate KV caches across engines~\cite{lin2024parrotefficientservingllmbased} to meet request SLOs and improve tail latencies. Overall, the above techniques optimize for \textit{independent LLM requests}, equivalent to a function-call in a general program. Instead, \text{\name} focuses on program-level optimizations, particularly scheduling—akin to how traditional OSs manage entire processes across CPU cores.

\subsection{Agentic Layer}
\label{sec:llm_agents}

\noindent \textbf{Agentic Programs.}
Above the LLM inference layer, developers build sophisticated \textit{agentic programs} to orchestrate interactions between agents, tools, and humans (Fig.~\ref{fig:agent_infra}). 
Specifically, this work focuses on LLM agents, defined as a tuple consisting of a system prompt specifying the agent's role and the LLM model class\footnote{LLM agents with identical system prompts but different models (e.g., LLaMA~\cite{llama}, Mistral~\cite{mistral}) are considered distinct\cite{mixtureofagents}.}. Similar to traditional OS processes and interrupts, agentic programs either interact directly with the LLM serving layer via LLM calls or engage in external interrupts—time spent outside an LLM engine. Specifically, agents can interact with tools to execute generic functions or external APIs, enabling control over environments such as databases, robotic systems, or the internet~\cite{schick2023toolformerlanguagemodelsteach, patil2023gorillalargelanguagemodel, yao2023webshopscalablerealworldweb,digirl,lmrobots,zhou2024llmenhanceddatamanagement}. Most importantly, agentic orchestration frameworks~\cite{openai_swarm}, such as LangChain~\cite{langchain, langgraph} and Autogen~\cite{wu2023autogenenablingnextgenllm}, provide developers with primitives to manage a program's control flow, determining when to execute agents, invoke tools, or request human input. Such primitives adhere to general programming semantics, including conditional statements, loops, error handling, and terminal conditions~\cite{zheng2024sglangefficientexecutionstructured, langgraph, wu2023autogenenablingnextgenllm, dspy}.
Finally, programs maintain a global history of outputs across agents, tools, and humans~\cite{lin2024parrotefficientservingllmbased,langgraph,memgpt,bufferofthoughts}. For instance, LLM-based chatbots accumulate messages between LLM agents' outputs and humans' inputs~\cite{openai2023gpt4}. Importantly, \text{\name} does not modify the program layer. Instead, it dynamically builds an internal state of the program’s execution graph (DAG) when the program runs, which is stored in a process table (\S\ref{sec:implementation}).

\begin{figure}[t]
\centering
\includegraphics[width=\linewidth]{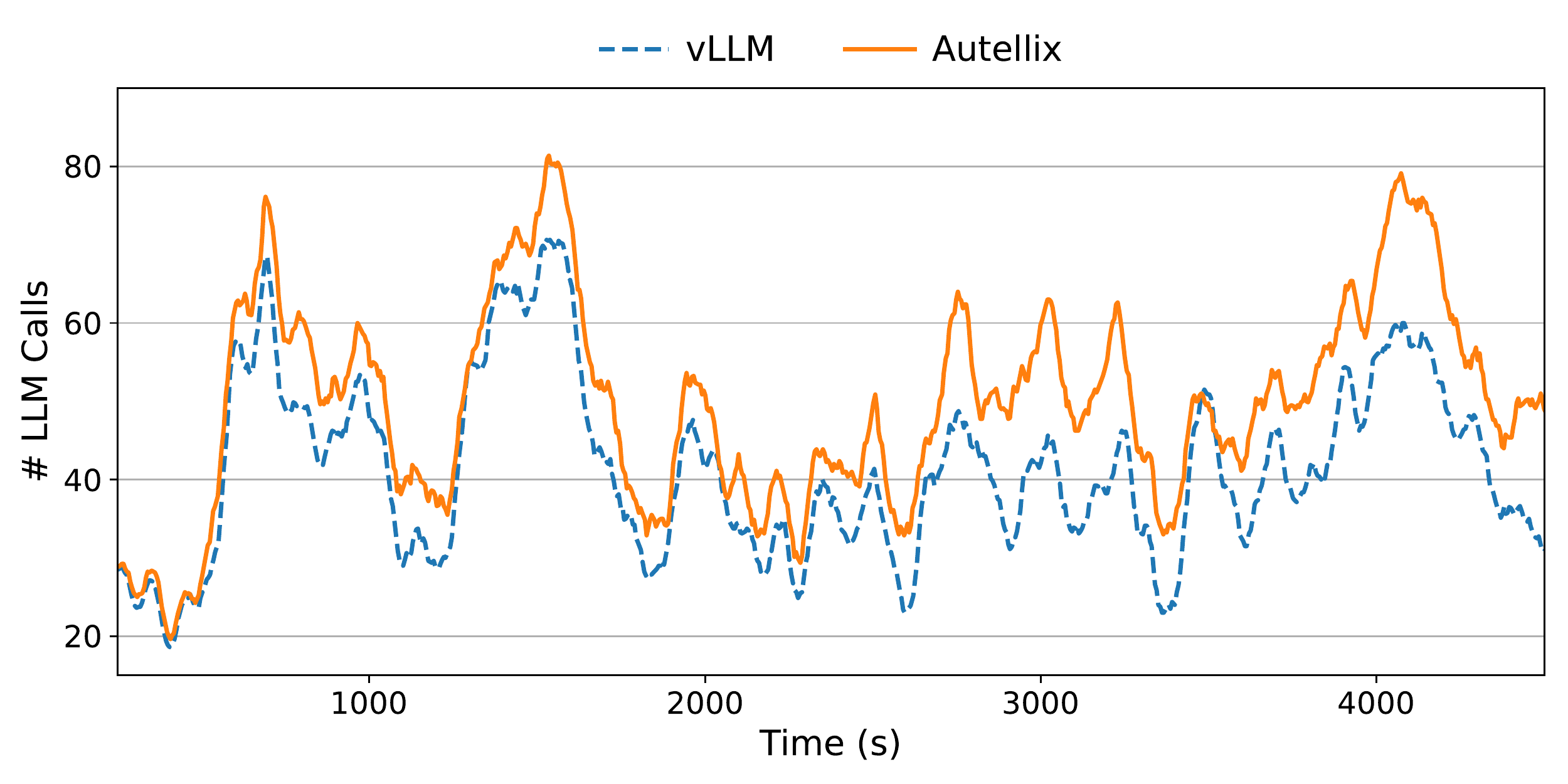}

\vspace{-2mm}
\caption{\small \textbf{Number of LLM calls in serving engine during steady state over 1 hour.}  Optimizing programs' wait times increases the volume of LLM calls at steady state.}
\vspace{-5mm}
\label{fig:steady_state}
\end{figure}
\vspace{2mm}
\noindent \textbf{Agentic Applications.} Beyond standard chatbots (Fig.~\ref{fig:chatbot_agent}), agentic applications, or instantiations of programs, automate or assist with complex tasks, including web or user-interface (UI) navigation (e.g. OpenAI's Operator)~\cite{OpenAIOperator, zhou2024webarenarealisticwebenvironment, gur2024realworldwebagentplanninglong,digirl}, resolving Github issues~\cite{jimenez2024swebench, yang2024sweagentagentcomputerinterfacesenable, wang2024openhandsopenplatformai}, solving IMO-level problems~\cite{deepmind_imo_silver_2024, kumarappan2024leanagentlifelonglearningformal}, fact-checking and summarizing claims from multiple sources (Fig.~\ref{fig:map_reduce_agent})~\cite{genspark2024, lin2024parrotefficientservingllmbased}, and enabling precise robotic control~\cite{rosser2024headsbetteronecollaborative}. Many applications scale inference time compute—the number of LLM calls and, correspondingly, total decode tokens—to improve their performance on complex tasks. These test-time methods include: step-by-step reasoning to decompose tasks~\cite{wei2023chainofthoughtpromptingelicitsreasoning, self-ask}, explicit thought injection to guide reasoning~\cite{yao2023reactsynergizingreasoningacting}, planning or searching to explore possible solutions~\cite{yao2023treethoughtsdeliberateproblem, Besta_2024, zhou2024languageagenttreesearch}, self-critique to evaluate actions~\cite{llmasjudge, stylus}, self-reflection to learn from failures~\cite{shinn2023reflexionlanguageagentsverbal, kumar2024traininglanguagemodelsselfcorrect}, and multi-agent collaboration~\cite{wu2023autogenenablingnextgenllm, societyofminds}. In particular, a single-threaded Reasoning and Acting (ReAct) agent, which combines chain-of-thought (CoT) techniques to efficiently act in an environment (Fig.\ref{fig:react_agent}), has recently been integrated on top of Deepseek-style (or o1-style) LLMs to enable automatic reasoning and tool calling~\cite{deepseekai2025deepseekr1incentivizingreasoningcapability, wang2024openhandsopenplatformai,RAGEN}. A multi-threaded program, Monte Carlo Tree Search (MCTS)~\cite{zhou2024languageagenttreesearch}, integrates parallel planning, self-critique, self-reflection, and multi-agent collaboration (Fig.~\ref{fig:mcts_agent}). Beyond MCTS, distributed programs may also incorporate best-of-N sampling, beam search, lookahead techniques, and genetic algorithms to explore and discover optimal solutions~\cite{chow2024inferenceawarefinetuningbestofnsampling, snell2024scalingllmtesttimecompute, davis2024networksnetworkscomplexityclass, lee2025evolvingdeeperllmthinking}. Given the probabilistic nature of LLMs, the breadth of inference-time techniques indicates that agentic programs and their applications exhibit three properties: (1) \textit{dynamic}, as different user prompts over the same program can yield entirely different execution patterns, (2) \textit{non-deterministic}, since the future is unknown, such as when a program decides to terminate, and (3) \textit{distributed}, with many programs leveraging parallel calls. Hence, \text{\name} is non-clairvoyant, operating with zero prior knowledge of programs' workloads or execution graphs.



\begin{figure}[t]
\centering
\includegraphics[width=\linewidth]{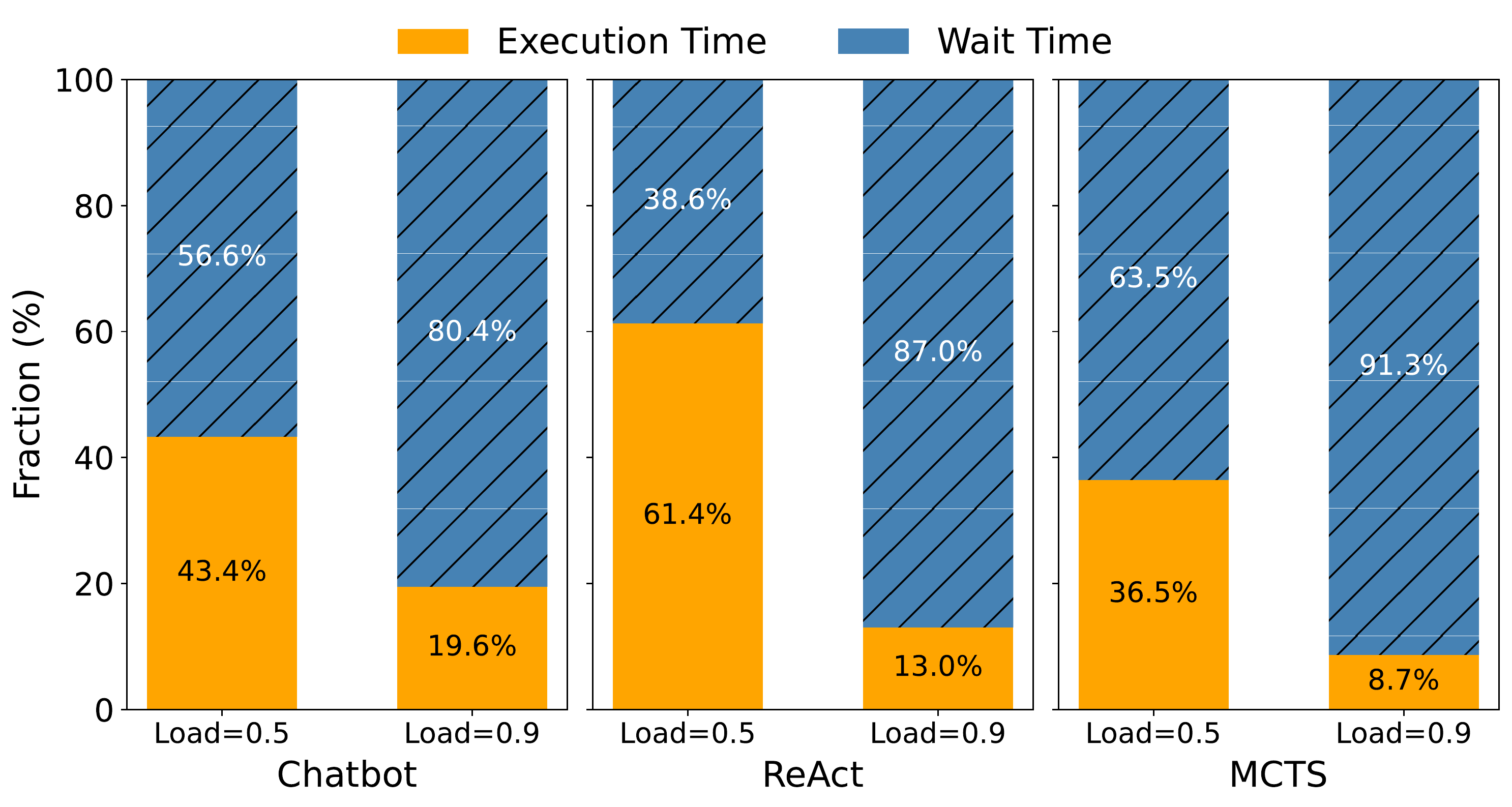}

\vspace{-2mm}
\caption{\small \textbf{Program execution and wait times, over different programs and system loads.} With moderate loads, programs spend the most time waiting. The duration of waiting depends on the workload.}
\vspace{-5mm}
\label{fig:high_load_wait_times}
\end{figure}

\section{Motivation}
\label{sec:motivation}
Today's AI agent infrastructure decouples LLM serving systems from agentic programs (\S\ref{sec:background}). As organizations shift from serving LLM queries to higher-level AI applications, LLM engines must optimize for program-level objectives, such as response times, or end-to-end latencies~\cite{lin2024parrotefficientservingllmbased}. Formally, a single-threaded program's end-to-end latency comprises three components: (1) \emph{waiting time}, the total queuing time of a program's LLM calls on the engine; (2) \emph{execution time}, the cumulative feedforward time of LLM calls; and (3) \emph{interceptions}, time spent waiting for external interrupts such as tool calls or human input. Since component (3) is unrelated to LLM serving, this section identifies problems and opportunities to reduce waiting (\S\ref{sec:wait_time_reduce}) and execution times (\S\ref{sec:execution_times_reduce}), subsequently addressed in the design of \text{\name}'s scheduling policies (\S\ref{sec:solution}).

\subsection{Program-level Wait Times}
\label{sec:wait_time_reduce}
\begin{figure}[t]
\centering
\begin{subfigure}[b]{0.49\columnwidth}
    \includegraphics[width=\linewidth]{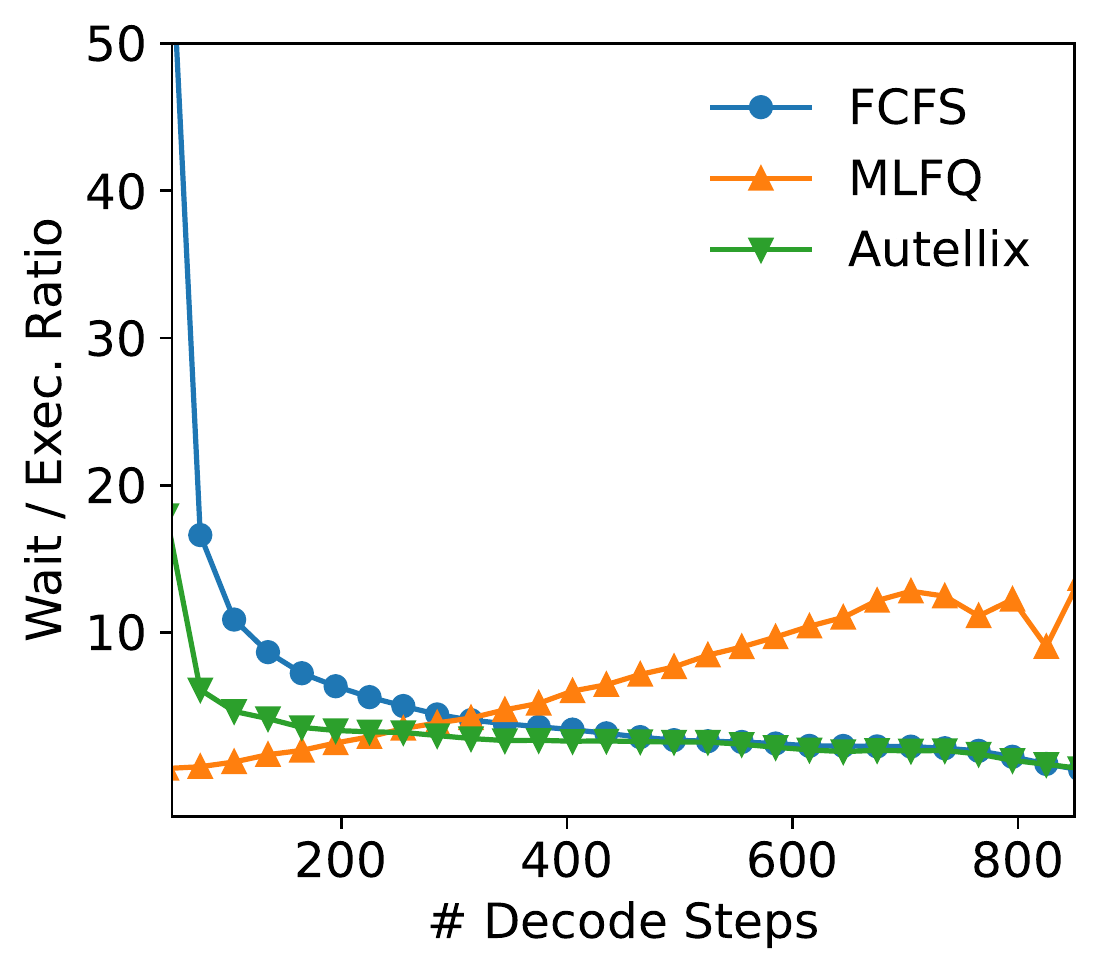}
    \caption{Chatbot, LLM Calls}
    \label{fig:sharegpt_call_block}
\end{subfigure}
\hfill
\begin{subfigure}[b]{0.49\columnwidth}
    \includegraphics[width=\linewidth]{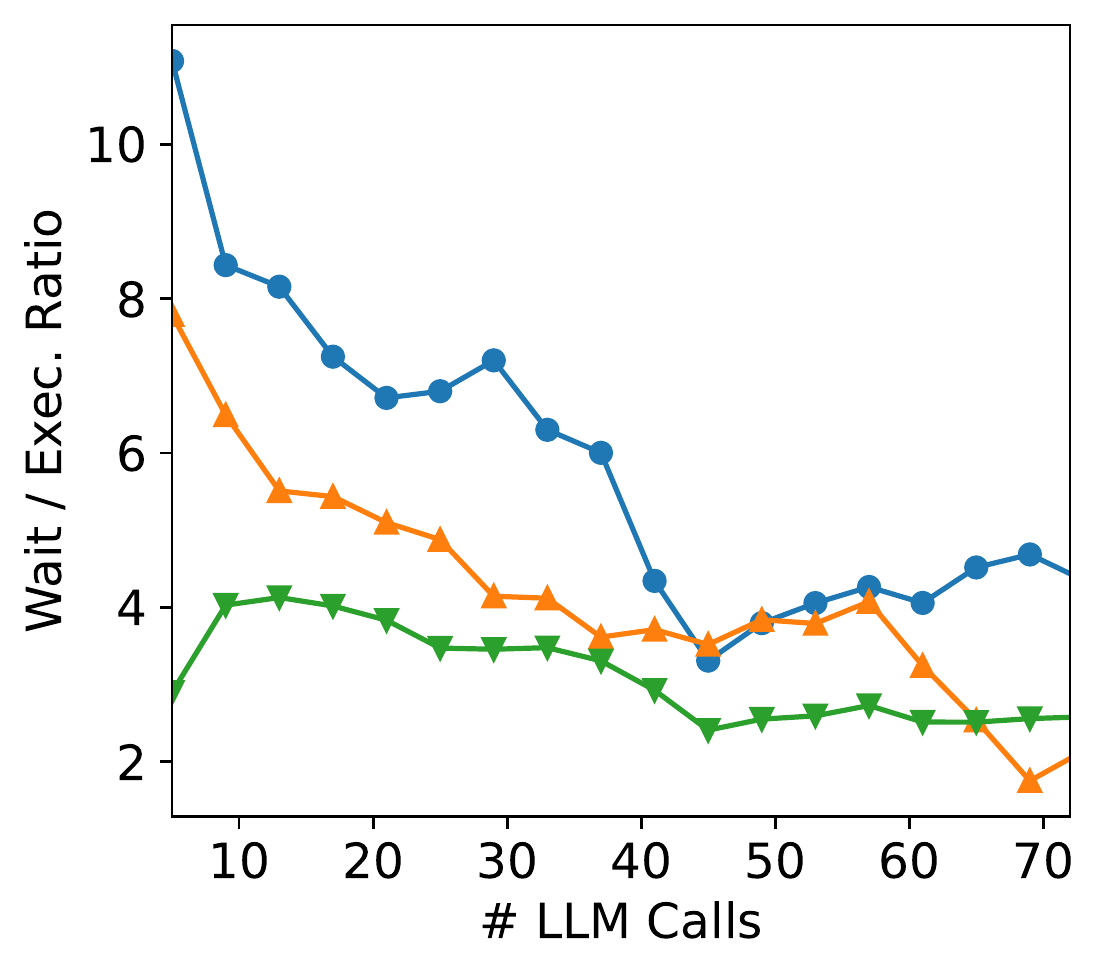}
    \caption{Chatbot, Programs}
    \label{fig:sharegpt_prog_block}
\end{subfigure}
\hfill
\vspace{2mm}
\begin{subfigure}[b]{0.49\columnwidth}
    \includegraphics[width=\linewidth]{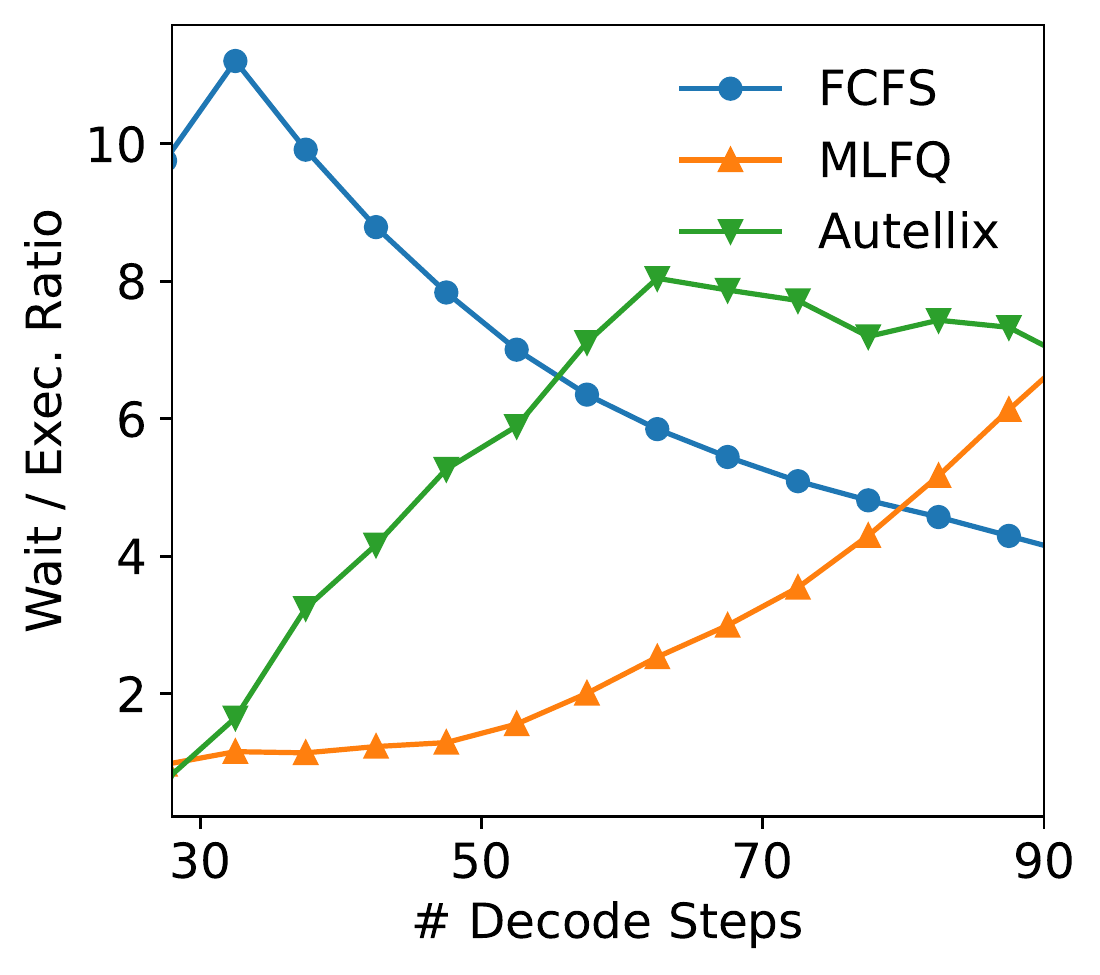}
    \caption{MCTS, LLM Calls}
    \label{fig:MCTS_call_block}
\end{subfigure}
\hfill
\begin{subfigure}[b]{0.49\columnwidth}
    \includegraphics[width=\linewidth]{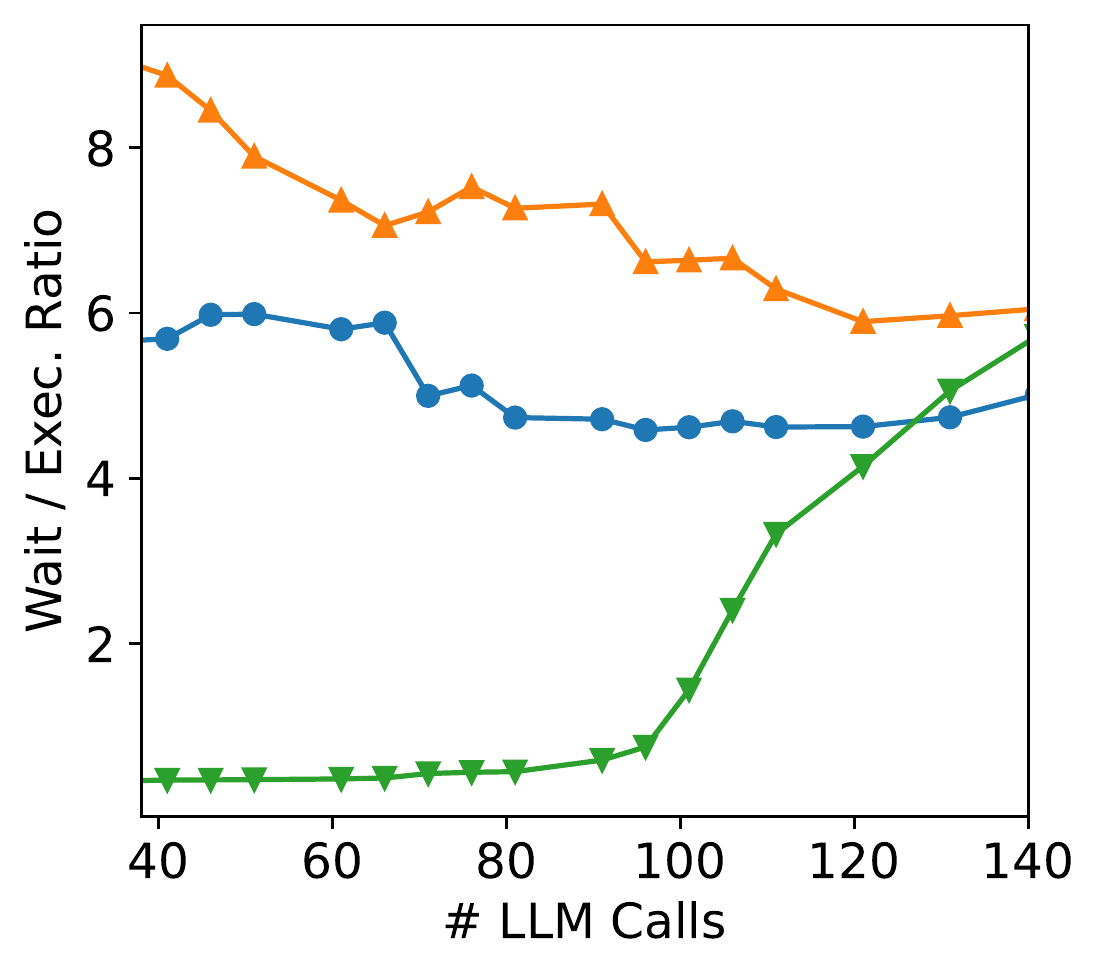}
    \caption{MCTS, Programs}
    \label{fig:MCTS_prog_block}
\end{subfigure}
\caption{\small \textbf{Ratio of Waiting to Execution Time for LLM Calls and Programs.} Head-of-line blocking occurs when short LLM calls and programs wait significantly longer than their execution times.}
\label{fig:hol}
\vspace{-5mm}
\end{figure}

Figure~\ref{fig:high_load_wait_times} shows that across various agentic workloads—from classic chatbots to ReAct and MCTS programs—the majority of a program's time is spent waiting as load increases. Hence, \text{\name} prioritizes reducing wait times, which not only improves program's latancies, but also increases LLM engine throughput. Faster call completions prompt programs to issue subsequent calls more quickly, increasing the arrival rate of LLM calls. Figure~\ref{fig:steady_state} illustrates steady-state behavior over a one-hour trace using LLaMA-3.1-8B~\cite{llama3} on a single A100-80GB GPU for entire chatbot conversations~\cite{sharegpt}. Compared to vLLM's first-come, first-served (FCFS) policy, \text{\name} consistently handles 10 additional concurrent LLM calls, offering more batching opportunities to improve throughput.

\vspace{2mm}
\noindent \textbf{Call-level Blocking.} The first challenge is LLM call-level \textit{head-of-line (HoL) blocking}. LLM calls with long decodes delay shorter ones, causing significant wait times~\cite{fastserve}. This issue is evident in serving engines like vLLM~\cite{vllm}, which wait for ongoing calls to finish decoding before scheduling new ones. HoL blocking is severe in our evaluated workloads with long-tailed distributions of decoding steps (Fig.~\ref{fig:trace-analysis}).

To measure blocking, Figure~\ref{fig:hol} measures the ratio of LLM requests' waiting time to execution time for Chatbot and MCTS workloads, as a function of output tokens. For FCFS policy, HoL blocking increases wait times for short LLM calls, increasing the ratio. Preemption, similar to how operating system schedulers interrupt long-running processes, mitigates HoL blocking by favoring shorter LLM calls. Figure~\ref{fig:hol} shows that Multi-Level Feedback Queue (MLFQ), a preemptive algorithm, leads to smaller ratios for short decodes. However, preemption without program-level statistics may not fully resolve the issue, as explained next.

\vspace{2mm}
\noindent \textbf{Program-level Blocking.} The second challenge is \emph{program-level HoL blocking}, where longer programs with many LLM requests delay shorter programs. Existing LLM schedulers are program-agnostic; they schedule individual LLM requests without considering their positions within the overall program, leading to suboptimal decisions. Our evaluation  shows a long-tailed distribution of LLM calls per program, which increases program-level blocking (\S\ref{sec:evaluation}).

To quantify program-level blocking, Figure~\ref{fig:hol} measures the ratio of programs' waiting time to execution time, with respect to number of LLM calls. For both workloads, FCFS and MLFQ incur higher ratios when the number of LLM calls is small, suggesting that short programs wait a long time. Due to this, preemptive scheduling policies, like MLFQ, may perform close to, or even worse, than FCFS (\S~\ref{sec:evaluation}). Without program-level context, MLFQ blindly prioritizes new LLM requests, leading to starvation of shorter programs when long programs' new LLM calls are prioritized.

\subsection{Program-level Execution Times}
\label{sec:execution_times_reduce}
\begin{figure}[t]
\centering
\begin{subfigure}[b]{0.49\columnwidth}
    \includegraphics[width=\linewidth]{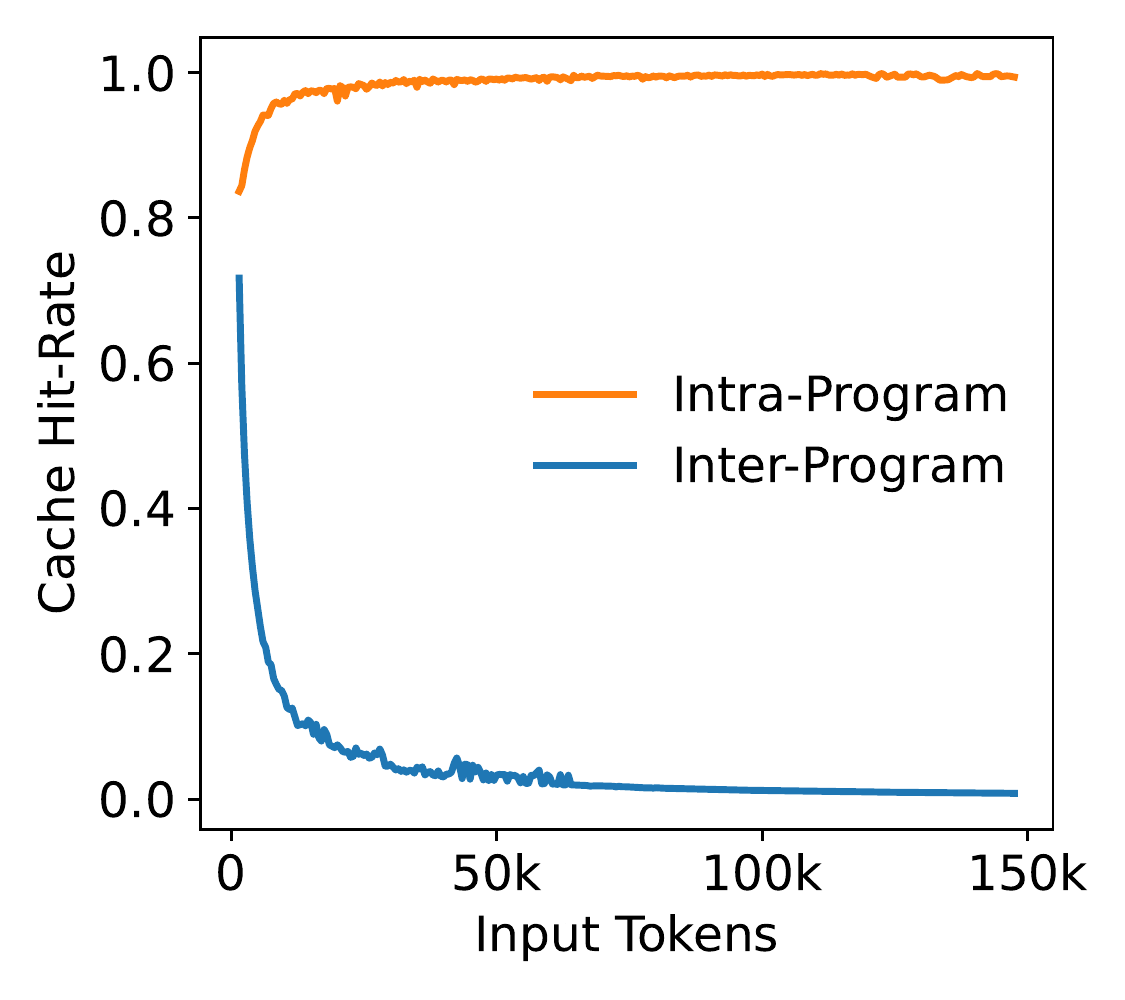}
    \caption{Single thread: Chatbot}
    \label{fig:sharegpt_prefix}
\end{subfigure}
\hfill
\begin{subfigure}[b]{0.49\columnwidth}
    \includegraphics[width=\linewidth]{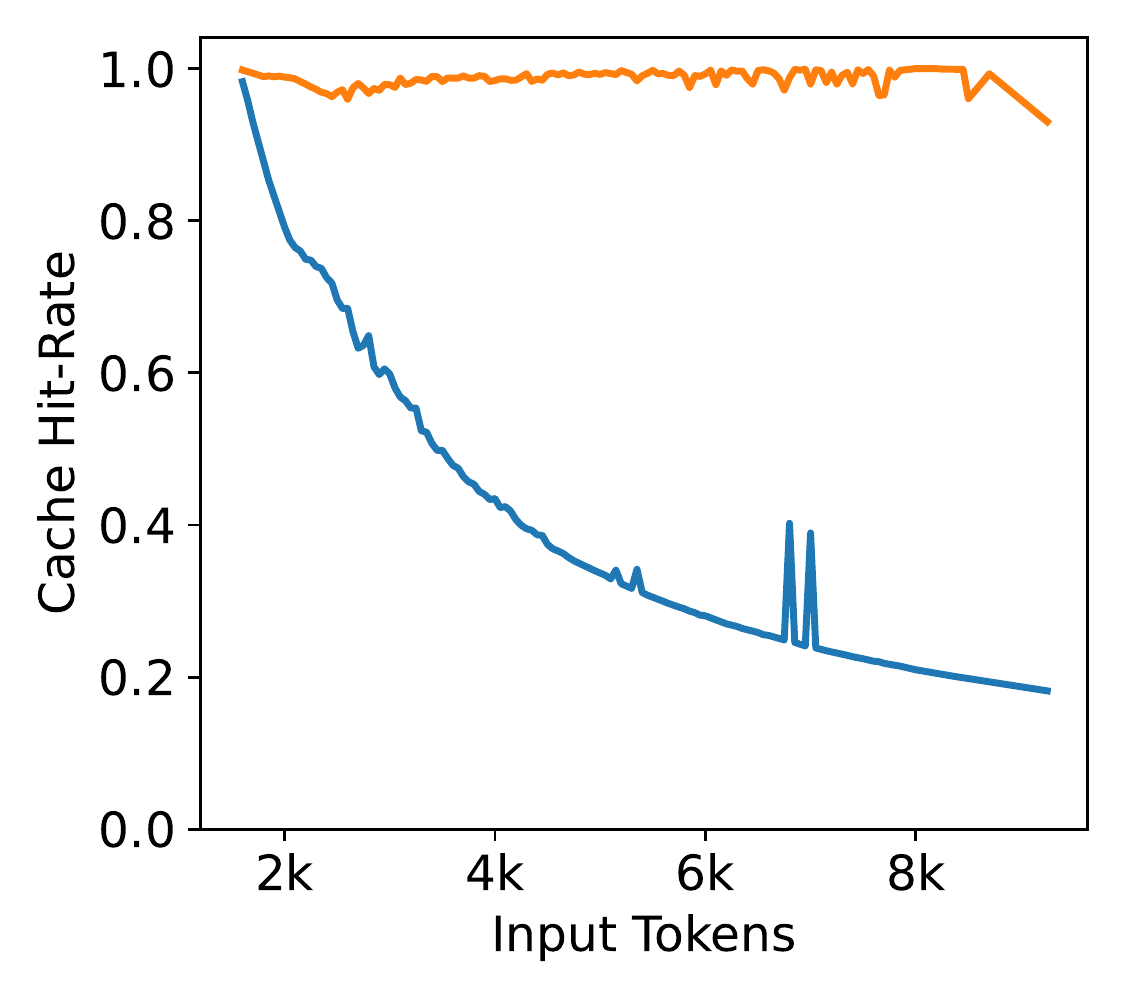}
    \caption{Multiple threads: MCTS}
    \label{fig:mcts_prefix}
\end{subfigure}
\caption{\small \textbf{Prefix cache hit rates for LLM calls within and across programs.} LLM calls within the same program often share KV cache, whereas LLM calls across programs typically do not.}
\label{fig:prefix_cache}
\vspace{-6mm}
\end{figure}

A program's execution time largely depends on how efficiently the LLM engine manages the prefill and decoding phases. In agentic workloads, which often feature long, cumulative prefills, \text{\name} focuses on optimizing prefill performance. Specifically, significant portions of prefill computation can be eliminated through prefix caching. This technique stores and reuses relevant key-value (KV) cache entries—such as the system prompt—across LLM requests~\cite{lin2024parrotefficientservingllmbased,zheng2024sglangefficientexecutionstructured}.

\vspace{2mm}
\noindent \textbf{Data Locality.} Figure~\ref{fig:prefix_cache} illustrates the average cache-hit rate as a function of input length. The cache-hit rate is defined as the percentage of precomputed input tokens in the LLM engine's KV cache for an incoming LLM call. Notably, within a single program, cache-hit rates remain above 90\% across all input lengths, indicating that LLM calls within the same program share identical contexts. In contrast, when considering different programs, the cache-hit rate decays exponentially with input length, suggesting that programs only share the system prompt. These results suggest that LLM serving systems across engines should consider a program's \textit{data locality}, as much of its KV cache can be reused for future LLM requests.

\section{\text{\name} Design}
\label{sec:solution}

We present \text{\name}'s overall architecture (\S\ref{sec:agentix_overview}) and then explore its two key components: (1) a program-aware scheduler (\S\ref{sec:agentix_policy}) designed to reduce both call-level and program-level blocking, and (2) a data locality–aware load balancer (\S\ref{sec:load_balancer}).

\subsection{Overview} \label{sec:agentix_overview}
\begin{figure}[t]
    \centering
    \includegraphics[width=\linewidth]{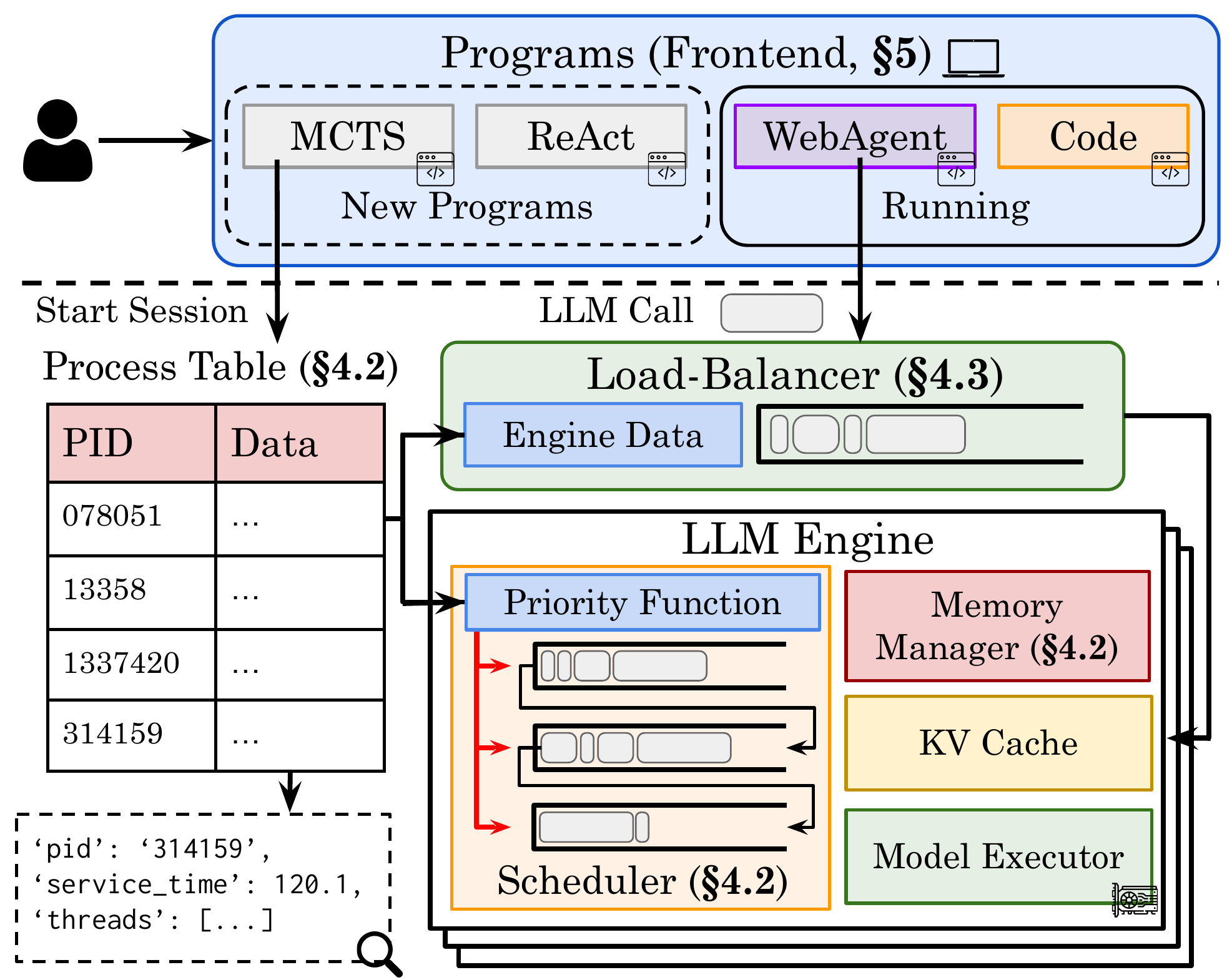}
    \caption{\small\textbf{\text{\name}'s system architecture.} Users run their programs locally, which initiates a stateful session and submits LLM calls to \text{\name}'s backend. \text{\name} leverages a global process table to track sessions and better inform its custom load-balancer and scheduler.}
    \label{fig:architecture}
     \vspace{-6mm}
 \end{figure}
\text{\name} is a higher-level serving engine designed for agentic programs rather than individual LLM requests.
\text{\name} focuses on three primary objectives: (1) improving overall program's end-to-end latency, for users, (2) maximizing GPU utilization for providers, and (3) mitigating program starvation to improve fairness, measured via 95th and 99th percentile latencies.


\vspace{1mm} \noindent\textbf{Assumptions.} \text{\name} is non-clairvoyant; it assumes
no knowledge of program arrivals, the structure of executed workflows, or general workload distributions. When a program arrives, its execution DAG is initially unknown; \text{\name} dynamically constructs an internal representation (IR) as the program runs. This flexibility enables \text{\name} to generalize to any program that invokes LLM calls on the underlying engine. While prior work~\cite{lin2024parrotefficientservingllmbased}
submits static LLM applications to the engine, \text{\name} assumes users run general Python programs on their local machines, which invoke \text{\name}'s backend (\S~\ref{sec:implementation}).

\vspace{1.5mm} 
\noindent \textbf{Architecture.} Figure~\ref{fig:architecture} illustrates \text{\name}'s overall architecture. Unlike existing LLM engines, which assumes LLM calls are stateless, \text{\name} is stateful: programs execute from the user's local machine, establish a session with the \text{\name}, and issue LLM calls over time with an associated session ID. We further detail the low-level implementation in Section~\ref{sec:implementation}. When a session starts, \text{\name} adds a corresponding entry to a global process table (\S\ref{sec:agentix_policy}). This table tracks program metadata, including total service time, thread-level metadata, and waiting times across programs' LLM calls. Both the engine-level scheduler (\S\ref{sec:agentix_policy}) and stateful load balancer (\S\ref{sec:load_balancer}) leverage the table to schedule LLM calls for the next decoding batch and route LLM calls to an engine based on their program's data locality. 

\subsection{Program-Aware Scheduler}
\label{sec:agentix_policy}
\begin{algorithm}[t]
    \small
    \caption{\text{\name}'s Program-Aware Scheduler}
    \label{algorithm:agentix}
    \begin{algorithmic}[1]

    \Procedure{Update\_Process\_Table}{\textbf{Call} $c$, \textbf{Table} $pt$}
        \State $\text{pd} = \text{pt}[\text{c.pid}]$ \label{alg:line1}
        \State \textcolor{cyan}{// Total service time (PLAS), max critical path (ATLAS)} \label{alg:line2}
        \State $\text{pd}\text{.service} = \max(\text{pd}\text{.service}, \text{c.service} + \text{c.model\_time})$ \label{alg:line3}
        \State \textcolor{cyan}{// Update other metrics...} \label{alg:line4}
        \State \ldots \label{alg:line5}
    \EndProcedure \label{alg:line6}
    \Procedure{Scheduler}{\textbf{Queues} $Q_{1}, \ldots, Q_{K}$, \textbf{Table} $pt$}
      \For{$\text{c} \in C_{arrived}$} \Comment{Arriving LLM calls} \label{alg:line9}
            \State \textcolor{cyan}{// Fetch priority with program ID} \label{alg:line10}
            \State $\text{c.service} = pt[\text{c.pid}]\text{.service}$ \label{alg:line11}
            \State $\text{c.q\_idx} = i, \text{ } s.t. \text{ } Q^{low}_{i} \leq \text{c.service} \leq Q^{hi}_{i}$ \label{alg:line12}
            \State $Q_{\text{c.q\_idx}}\text{.append(c)}$, $\text{c.quanta} = Q_{\text{c.q\_idx}}\text{.quanta}$ \label{alg:line13}
      \EndFor \label{alg:line14}

      \For{$\text{c} \in \{Q_{1}, Q_{2}, ..., Q_{K}\}$} \label{alg:line16}
        \If{$\text{c.finished()}$} \Comment{Finished jobs update table} \label{alg:line17}
            \State \textsc{Update\_Process\_Table}$(\text{c}, \text{ }$pt$)$ \label{alg:line18}
            \State $Q_{\text{c.q\_idx}}\text{.remove(c)}$ \label{alg:line19}
        \EndIf \label{alg:line20}

        \If{$\text{c.quanta} \leq 0$} \Comment{Call demotion} \label{alg:line21}
            \State $Q_{\text{c.q\_idx}}\text{.remove(c)}$, $Q_{\text{c.q\_idx}+1}\text{.append(c)}$ \label{alg:line22}
            \State $\text{c.q\_idx} += 1$, $\text{c.quanta} = Q_{\text{c.q\_idx}}\text{.quanta}$ \label{alg:line23}
        \EndIf \label{alg:line24}

        \State $\text{wait} = pt[\text{c.pid}]\text{.wait} + \text{c.wait}$ \label{alg:line25}
        \State $\text{service} = pt[\text{c.pid}]\text{.service} + \text{c.model\_time}$ \label{alg:line26}
        \If{$\text{wait}/\text{service} \geq \beta$} \Comment{Anti-Starvation} \label{alg:line27}
            \State $Q_{\text{c.q\_idx}}\text{.remove(c)}$, $Q_{1}\text{.append(c)}$ \label{alg:line28}
            \State \textcolor{cyan}{// Reset waiting and model execution times} \label{alg:line29}
            \State $\text{c.wait}=0$, $\text{c.model\_time}=0$ \label{alg:unknown}
        \EndIf \label{alg:line31}
      \EndFor \label{alg:line32}

      \State $B_{out} = []$ \Comment{Schedule next batch of LLM calls} \label{alg:line33}
      \For{$\text{c} \in \{Q_{1}, Q_{2}, ..., Q_{K}\}$} \label{alg:line34}
        \If{$\text{engine.can\_fit(c)}$} \label{alg:line35}
            \State $B_{out}\text{.append(c)}$ \label{alg:line36}
        \Else \label{alg:line37}
            \State \textbf{break} \label{alg:line38}
        \EndIf \label{alg:line39}
      \EndFor \label{alg:line40}
    \EndProcedure \label{alg:line41}
    \end{algorithmic}
\end{algorithm}

We present a general, efficient scheduler designed to minimize programs' response times, or end-to-end latencies, without a-priori knowledge. To mitigate head-of-line blocking at both the program and call levels, \text{\name} assigns priorities to calls based on program-level statistics (e.g., total accumulated runtime, \S\ref{sec:program-priority}) and dynamically preempts calls (\S\ref{sec:preemption}). The complete scheduling algorithm is shown in Alg.~\ref{algorithm:agentix}.

\subsubsection{Program-level Prioritization}
To implement program-level prioritization effectively, \text{\name} relies on a global process table that tracks essential program metrics, enabling more informed scheduling decisions across both single- and multi-threaded programs.
\label{sec:program-priority}

\vspace{1.5mm}
\noindent \textbf{Process Table.} Inspired by traditional operating systems, \text{\name} maintains a global process table that records the state of all running programs. When a new program arrives, \text{\name} adds a corresponding entry; when the program completes, this entry is removed. Each program entry in the process table tracks the following metrics:
\begin{itemize}[itemsep=0pt, parsep=0pt, topsep=0pt, partopsep=0pt, leftmargin=*]
    \item \textit{Service time:} For single-threaded programs, this is the cumulative execution time of all completed calls on the LLM engine's model executor. For multi-threaded programs, it is the longest observed critical path's execution time.
    \item \textit{Waiting time:} The time spent in the LLM engine's scheduler queue—used for anti-starvation.
    \item \textit{Engine ID(s):} The engine(s) that the program is currently running on—used for \text{\name}'s load-balancer. (\S~\ref{sec:load_balancer}).
    \item \textit{Threads Metadata:} Each thread corresponds to an active LLM call. Hence, we keep track of a program's active LLM calls and their individual arrival, waiting, and service times. 
    \item \textit{Most recent call arrival:} The last time a new LLM call arrived for this program—used for tracking stale programs.
    \item \textit{Most recent call completion:} The last time an LLM call finished—used for detecting long external interrupts.
\end{itemize}
When a program's LLM call completes, the table is updated accordingly. With the process table, the scheduler can reason about the global state of each program to schedule LLM calls.

\vspace{1.5mm}
\noindent \textbf{Single-Threaded Programs.}
\begin{figure}[t]
\centering
\includegraphics[width=\linewidth]{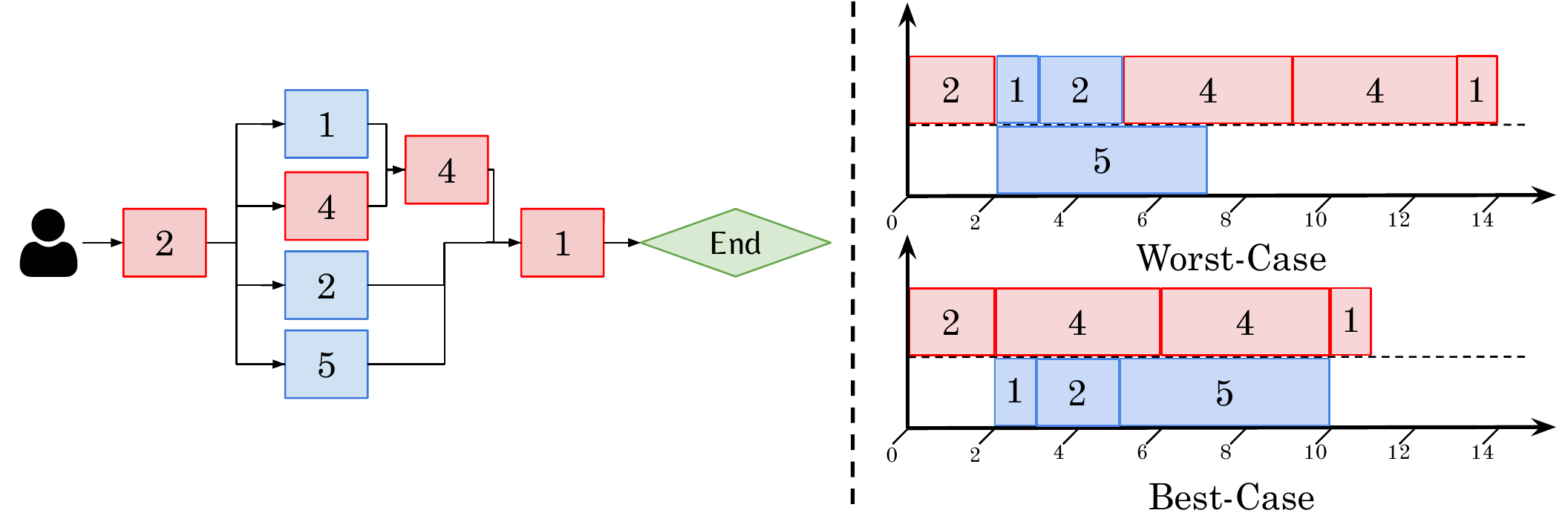}
\vspace{-2mm}
\caption{\small \textbf{Critical path for multi-threaded programs.} (Left) Example of a critical path through a DAG. (Right) Best-case scenario makespan, 14 units, versus worst-case makespan. 11 units.}
\label{fig:dag_scheduling}
\vspace{-5mm}
\end{figure}
Scheduling policies like Shortest-Job-First (SJF) and Shortest-Remaining-Processing-Time (SRPT) minimize response times optimally in single- and multi-server settings~\cite{srpt_optimal, sjf_optimal}. However, these require exact knowledge of program runtimes, violating \text{\name}'s non-clairvoyance assumption. Instead, the Least-Attained-Service (LAS) algorithm~\cite{las}, widely used in information-agnostic settings such as data center networking~\cite{las_datacenter, coflow_dag} and deep learning clusters~\cite{tiresias}, offers a practical alternative.

We introduce \textit{Program-Level Attained Service}, or \textit{PLAS}, extending LAS to programs. For a single-threaded program, its service time is the total runtime of all prior completed LLM calls. Formally, if the $j$th LLM call $c_j$ with program ID of $c_{j}.id$ is submitted, \textit{PLAS} assigns a priority $p(c_j)$ to $c_j$ based on the sum of all runtimes, $t_k$, of all prior LLM calls with the same ID:

\vspace{-2mm}
\begin{equation}
\label{equation:plas}
p(c_j) = \sum_{\substack{k < j \\ c_{k}.id = c_{i}.id}} t_k
\vspace{-2mm}
\end{equation}

\noindent Here, large priority values mean lower priority. To reduce computation, the scheduler reads the program’s total service time from the process table (Line~\ref{alg:line11}). When an LLM call completes, its program’s total service time is updated (Line~\ref{alg:line3}). Thus, \textit{PLAS} naturally favors calls from programs that have received less total service, helping shorter programs finish earlier and reducing response times.

\vspace{1.5mm}
\noindent \textbf{Multi-Threaded Programs.} Unlike single-threaded programs, multi-threaded programs are modeled as dynamic DAGs of LLM calls. Unfortunately, a program’s completion time is dictated by the DAG's \emph{critical path}—the longest sequence of dependent calls from start to finish, illustrated in Figure~\ref{fig:dag_scheduling}. No matter how many parallel LLM calls an engine can process, the program only terminates when all calls along the critical path have finished. Furthermore, without considering critical paths, schedulers achieve sub-optimal completion times for programs; in Figure~\ref{fig:dag_scheduling}, the DAG's makespan increases from 11 to 14 units. 

To address this, we introduce \textit{Adaptive Thread-Level Attained Service (ATLAS)}, a pragmatic generalization of \textit{PLAS}, that prioritizes calls based on their service times along their programs' critical paths. \textit{ATLAS} aims to assign each newly arrived call \( c_j \) a priority \( p(c_j) \) based on the priorities and completed service times of its parents \(\mathcal{P}(c_j)\) in the same program:

\vspace{-5mm}
\begin{equation}
\label{equation:atlas}
p(c_j) = 
\begin{cases} 
0 & \text{if } c_j \text{ is root} \\
\max_{c_k \in \mathcal{P}(c_j)} \{ p(c_k) + t_k \} & \text{otherwise}
\end{cases}
\end{equation}

\noindent Here, \( t_k \) is the execution time of a parent call \( c_k \). By recursively combining parent priorities and runtimes, \( p(c_j) \) estimates the longest chain of accumulated service time leading to \( c_j \), providing a non-clairvoyant estimation of the critical path.

However, achieving both objectives—favoring short programs while also prioritizing the longest, critical-path threads—is nontrivial. To solve this, \textit{ATLAS} maintains a single scalar per program in its process table: the longest observed critical path.  Each active LLM call in a program inherits this value as its initial priority, and upon call completion, updates the scalar only if its own critical path is longer (Line~\ref{alg:line3}). This simple mechanism continuously refines the program’s critical path estimate without tracking dependencies between LLM calls. Consequently, \textit{ATLAS} favors programs and LLM calls with shorter critical paths, effectively approximating a Least-Attained-Service policy for dynamic DAGs. Furthermore, as all calls of a given program derive their priorities from the same entry, the scheduler naturally groups a program’s parallel calls, preventing straggler threads from delaying programs' completion.

\subsubsection{Preemptive Scheduling}
\begin{figure}[t]
    \centering
    \includegraphics[width=\linewidth]{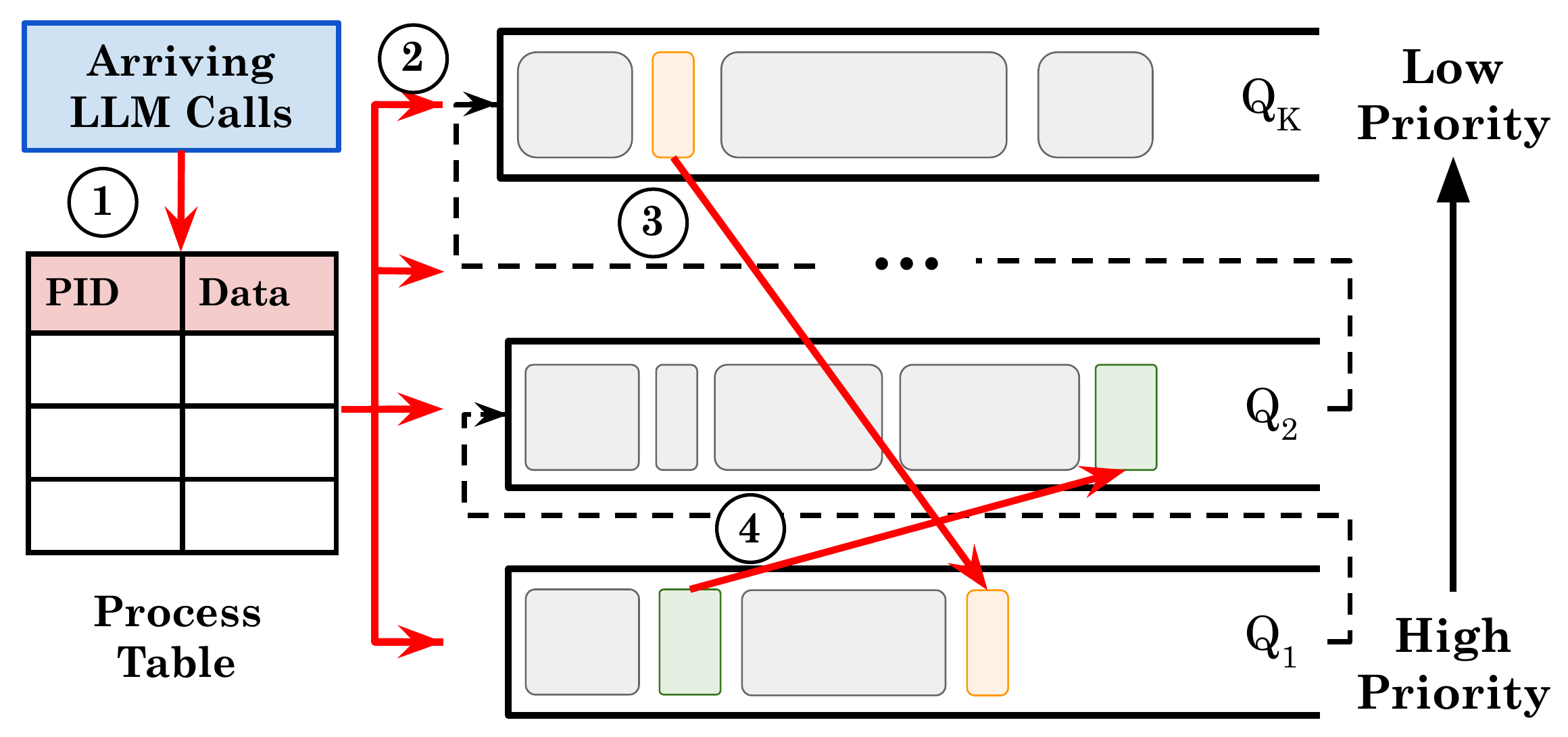}
    \caption{\small\textbf{LLM call lifecycle based on discretized prioritization.}}
    \label{fig:agentix_mlfq}
     \vspace{-3mm}
 \end{figure}
\label{sec:preemption}
\text{\name} assigns priorities to each LLM call based on their program's history. However, scheduling and preempting programs based on continuous priorities can degrade into worst-case round-robin scheduling~\cite{coflow_dag}, which performs worse than FCFS, and incur unnecessary context switches, including frequent KV-cache swaps between CPU and GPU~\cite{fastserve}. To avoid this, \text{\name} discretizes priorities into a finite set of queues, akin to multi-level feedback queues (MLFQ) in operating systems~\cite{coflow_dag, tiresias, arps_ostep_mlfq}.

\vspace{1.5mm}
\noindent \textbf{Multi-level Program-based Scheduling} \text{\name} bins and discretizes LLM calls' priorities into K queues ($Q_1, Q_2, \dots, Q_K$), where priorities decrease from $Q_1$ to $Q_K$. Each queue $Q_i$ covers a priority range $[Q_i^{lo}, Q_i^{hi})$, with $Q_1^{lo}=0$, $Q_K^{hi}=\infty$, and $Q_{i+1}^{lo}=Q_i^{hi}$.

In Figure~\ref{fig:agentix_mlfq}, when an LLM call arrives, \text{\name} looks up its program’s priority $p(c)$, based on the process table (\ding{172}, Line~\ref{alg:line11}). Unlike traditional MLFQ, where new calls all start at the highest priority queue $Q_1$, LLM calls are assigned to the $i$th queue based on discretized priorities, $p(c)\in[Q_i^{lo}, Q_i^{hi})$ (\ding{173}, Line~\ref{alg:line12}). Subsequently, calls receive the queue's time quantum and execute in FCFS order within their queue (Line~\ref{alg:line13},~\ref{alg:line36}).
Once a call exhausts its quantum, it is demoted to a lower priority queue (\ding{174}, Lines~\ref{alg:line21}-\ref{alg:line24}). If the call waits too long, \text{\name} employs anti-starvation mechanisms, described next (\ding{175}, Lines~\ref{alg:line25}-\ref{alg:line31}). Finally, when a call completes decoding, it updates the process table (Lines~\ref{alg:line17}-\ref{alg:line19}).

\vspace{1mm}
\noindent \textbf{Anti-Starvation.} Discrete prioritization, or MLFQ-style algorithms, incurs the starvation of long, low-priority programs~\cite{tiresias, fastserve, coflow_dag}. Simple anti-starvation techniques—such as promoting calls that have waited past a threshold—reduces \text{\name} to naive MLFQ, where long program's LLM calls, which are now in $Q_1$, interrupt short programs~\cite{fastserve, arps_ostep_mlfq}. Hence, we also utilize the process table to measure program-level starvation. Concretely, for a program $p$, \text{\name} promotes call $c$ to $Q_1$ if the ratio of total waiting time ($W_\text{total} = W_p + W_c$) to service time ($T_\text{total} = T_p + T_c$) exceeds a threshold $\beta$:
\vspace{-2mm}
\[ \frac{W_{\text{total}}}{T_{\text{total}}} \geq \beta \]
\vspace{-3mm}

\noindent Varying $\beta$ presents a trade off between programs' average response times and fairness. After promotion, only $W_c$ and $T_c$, or the calls' wait and run time, are set to zero, to ensure programs' threads, or active LLM calls, are likely all promoted together (Line~\ref{alg:unknown}). 

\vspace{1mm}
\noindent \textbf{Memory Management.} With preemptive scheduling, LLM engines must handle a large volume of concurrent LLM calls, leading to frequent GPU-CPU transfers as KV-cache blocks are repeatedly swapped to serve different requests~\cite{fastserve}. Prior work mitigates this swapping overhead by proactively swapping KV-cache for the next iteration of LLM requests while processing the current ones~\cite{fastserve}. However, \text{\name} is synchronous and requires real-time updates for each call's time quantum and the process table. Instead, \text{\name} employs two key optimizations to reduce both the frequency and overhead of GPU-CPU swapping respectively.

First, \text{\name} reduces total swaps by adopting multi-step scheduling, running the scheduler once every $N$ decoding steps rather than at every step. As some requests may complete early, our scheduler overprovisions queued requests already on the GPU, ensuring that new requests are immediately added when some requests finish before $N$ steps. Second, \text{\name} employs a more efficient GPU-CPU swap kernel. Instead of calling separate asynchronous transfers for each block, our kernel gathers all KV blocks into a contiguous buffer and transfers them in one operation—increasing PCIe bandwidth by reducing fragmentation, reducing per-block overhead, and lowering end-to-end swap latency (\S\ref{sec:implementation}).

\subsection{Load Balancer}
\label{sec:load_balancer}


\begin{algorithm}[t]
    \small
    \caption{\text{\name}'s Load Balancer}
    \label{algorithm:load_balancer}
    \begin{algorithmic}[1]

    \Procedure{Load\_Balancer}{\textbf{Call} $\text{c}$, \textbf{Table} $pt$, \textbf{List} $\text{Engines}$}
        \If{$\Call{Len}{\text{c.tokens}} \leq 2048$} \Comment{Small request} 
            \State $\text{assigned\_engine} = \Call{Least\_Used}{\text{Engines}}$ 
        \Else
            \If{$\text{c.pid} \in pt$} \Comment{Program already assigned to engine} 
                \State $\text{assigned\_engine} = pt[\text{c.pid}]$ 
            \Else
                \State \textcolor{cyan}{// Select the least utilized engine} 
                \State $\text{assigned\_engine} = \Call{Least\_Used}{\text{Engines}}$ 
                \State $pt[\text{c.pid}] = \text{assigned\_engine}$ 
            \EndIf
        \EndIf
    
        \State \Return $\text{assigned\_engine}$
    \EndProcedure
    
    \Procedure{Least\_Used}{\textbf{List} $\text{Engines}$}
        \State \textcolor{cyan}{// Query engine workloads in parallel} 
        \State $\text{workloads} = \Call{Query\_Engine\_Workloads}{\text{Engines}}$
        \State $\text{least\_used\_engine} = \Call{Argmin}{\text{workloads}}$ 
        \State \Return $\text{least\_used\_engine}$
    \EndProcedure
    \end{algorithmic}
\end{algorithm}
As agentic workloads scale, deploying multiple engine replicas is necessary. However, distributing requests without considering data locality yields suboptimal performance~\cite{preble}.

Our analysis for agentic workloads (§\ref{sec:execution_times_reduce}) highlights a critical distinction between short and long requests. Short requests below 2048 tokens achieve high cache hit rates ($\geq75\%$) across any engine, due to common system prompts. Enforcing data locality for these requests offers negligible gains and risks skewing engine utilization when large, parallel programs dominate specific engines. Thus, simply balancing short requests across the least-loaded engines preserves performance with minimal overhead. Conversely, longer requests are far more sensitive to their programs' data locality. Their substantial prefix overlap with a given program significantly reduces recomputation when consistently routed to the same engine, justifying occasional queuing delays.

While prior work relies on complex prefix trees to quantify data locality~\cite{preble}, our simple method dynamically routes short requests to the least-loaded engine and pins longer requests to their programs’ corresponding engines. Algorithm~\ref{algorithm:load_balancer} formalizes this approach, and our evaluation shows that \text{\name}’s load balancer improves both throughput and latency across heterogeneous workloads (\S\ref{sec:evaluation}).

\section{Implementation}
\label{sec:implementation}
\text{\name} is a multi-engine LLM inference serving system comprising a frontend, scheduler, and load balancer—totaling 5k lines of Python and CUDA/C++ code.

\vspace{1mm}
\noindent \textbf{Frontend.} \text{\name}’s frontend extends OpenAI’s Chat Completion and vLLM’s Python APIs~\cite{vllm, chat_completions} to provide a stateful interface that appears stateless to developers. Users simply import \text{\name}’s library into their Python applications, and upon program initialization, \text{\name} automatically issues a \texttt{start\_session} request to the backend. This operation returns a unique session identifier and creates a corresponding entry in the process table. Subsequent LLM calls are transparently annotated with the appropriate session, program, and thread IDs before being dispatched to the backend. When the program completes or encounters an error, \text{\name} invokes \texttt{end\_session}, removing the associated entry from the process table. As a research prototype, the current frontend lacks safeguards against user modification of the underlying package; addressing this limitation remains future work.


\vspace{1mm}
\noindent \textbf{LLM Engine.} 
\text{\name} builds on vLLM v0.6.1~\cite{vllm}. To keep changes localized, we modify only the scheduler by integrating new policies (\textit{PLAS}, \textit{ATLAS}, and MLFQ) and memory swapping kernels for efficiency. This ensures straightforward experimentation and clear attribution of performance gains. The scheduler follows the algorithm described in the previous section (\S\ref{sec:solution}). We've also noticed in vLLM, each Key-Value (KV) block is transferred individually via \texttt{cudaMemcpyAsync}, creating small fragmented transfers that underutilize PCIe bandwidth and incur high overhead such as repeated DMA setups. To address this, we allocate a host buffer and consolidate all KV blocks into a single contiguous chunk, enabling one bulk transfer. The results are shown in the next section (\S\ref{sec:evaluation}).

\vspace{1mm}
\noindent \textbf{Multi-engine.} vLLM currently lacks the ability to manage multiple LLM engines at the same time. To better evaluate our load balancing strategy, we built \texttt{AsyncMultiLLMEngine} atop of \texttt{AsyncLLMEngine}. Each LLM engine replica runs in a dedicated Python process, and a coordinating meta-engine manages these replicas via standard inter-process communication (IPC) primitives such as \texttt{mp.Queue} and \texttt{mp.Pipe}. When the meta-engine receives a request, it assigns the request to the appropriate replica, returning a future-like object to the frontend without blocking. The selected engine process executes the task asynchronously and sends the completed result back through the IPC channel. Upon receiving the result, the meta-engine resolves the future and provides the output to the frontend. This design allows multiple requests to be processed in parallel, with the meta-engine acting as a non-blocking coordinator that handles routing, resource assignment, and result collection.



\section{Evaluation}
\label{sec:evaluation}
In this section, we analyze representative agentic workloads, evaluate \text{\name}'s performance against state-of-the-art LLM serving systems, and ablate its design choices.

\subsection{Workloads}
\begin{figure}[t]
\centering
\begin{subfigure}[b]{0.49\columnwidth}
    \includegraphics[width=\linewidth]{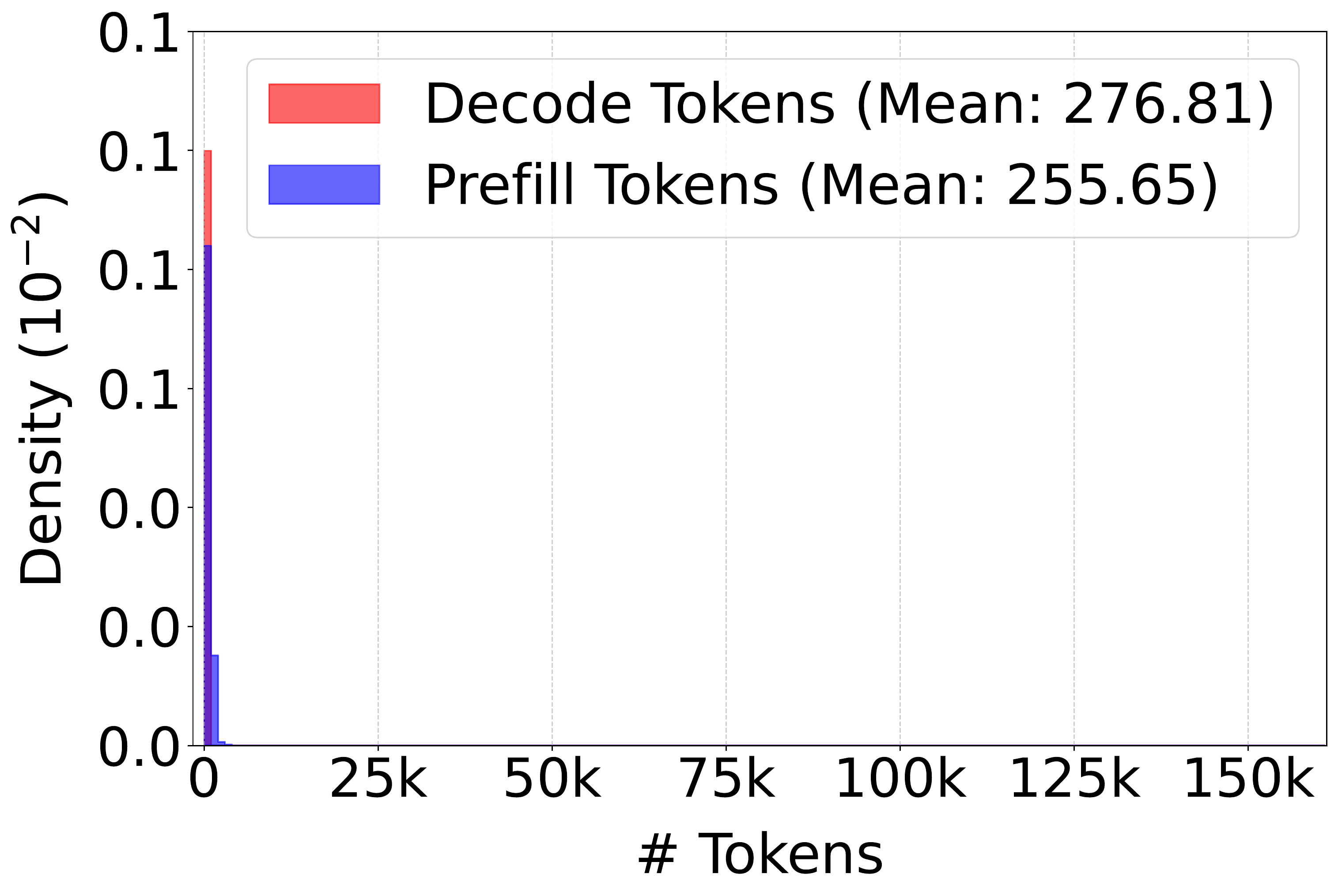}
    \caption{ShareGPT}
    \label{fig:sharegpt-llm-call-analysis}
\end{subfigure}
\hfill
\begin{subfigure}[b]{0.49\columnwidth}
    \includegraphics[width=\linewidth]
    {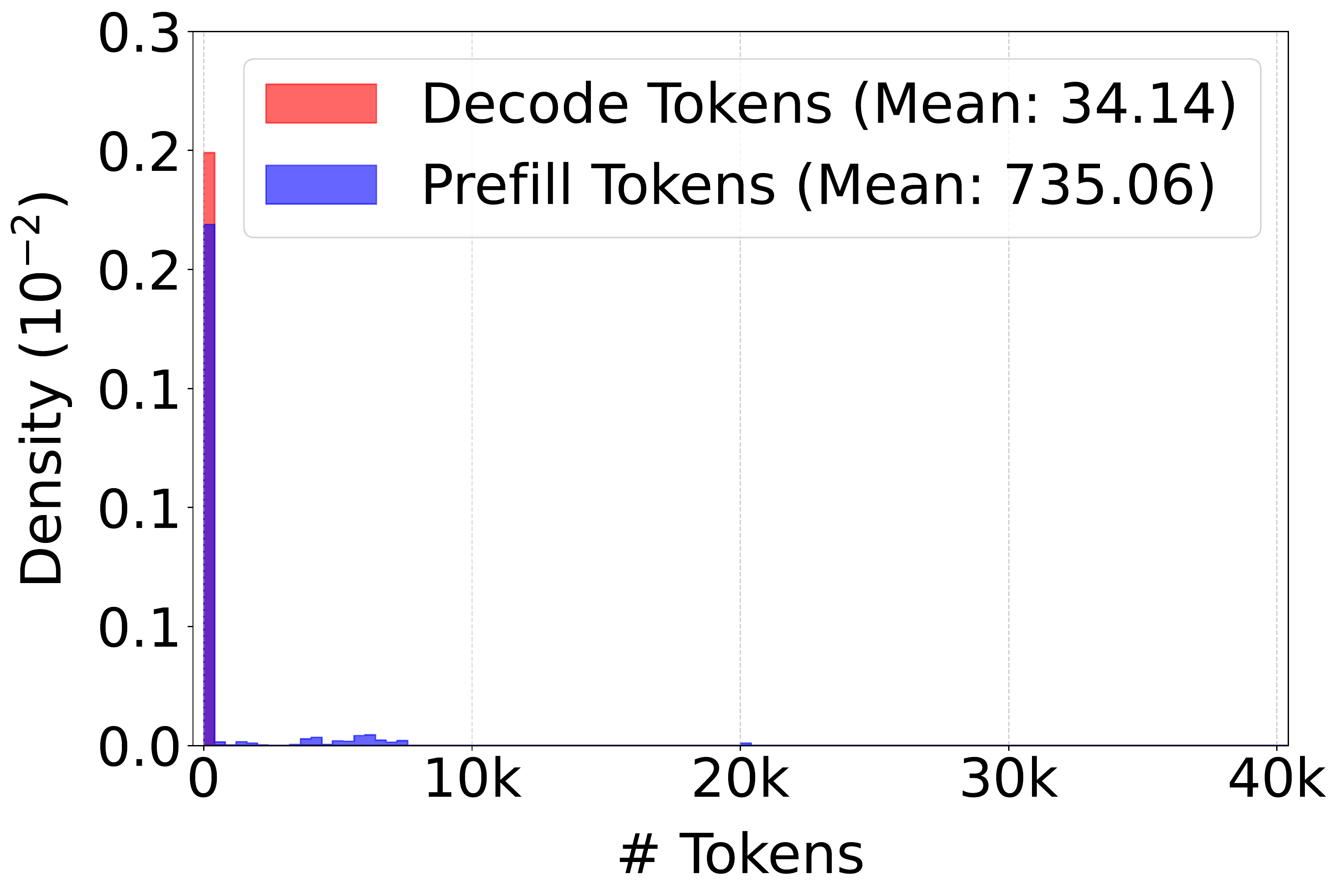}
    \caption{BFCL}
    \label{fig:bfcl-llm-call-analysis}
\end{subfigure}
\vskip\baselineskip
\begin{subfigure}[b]{0.49\columnwidth}
    \includegraphics[width=\linewidth]
    {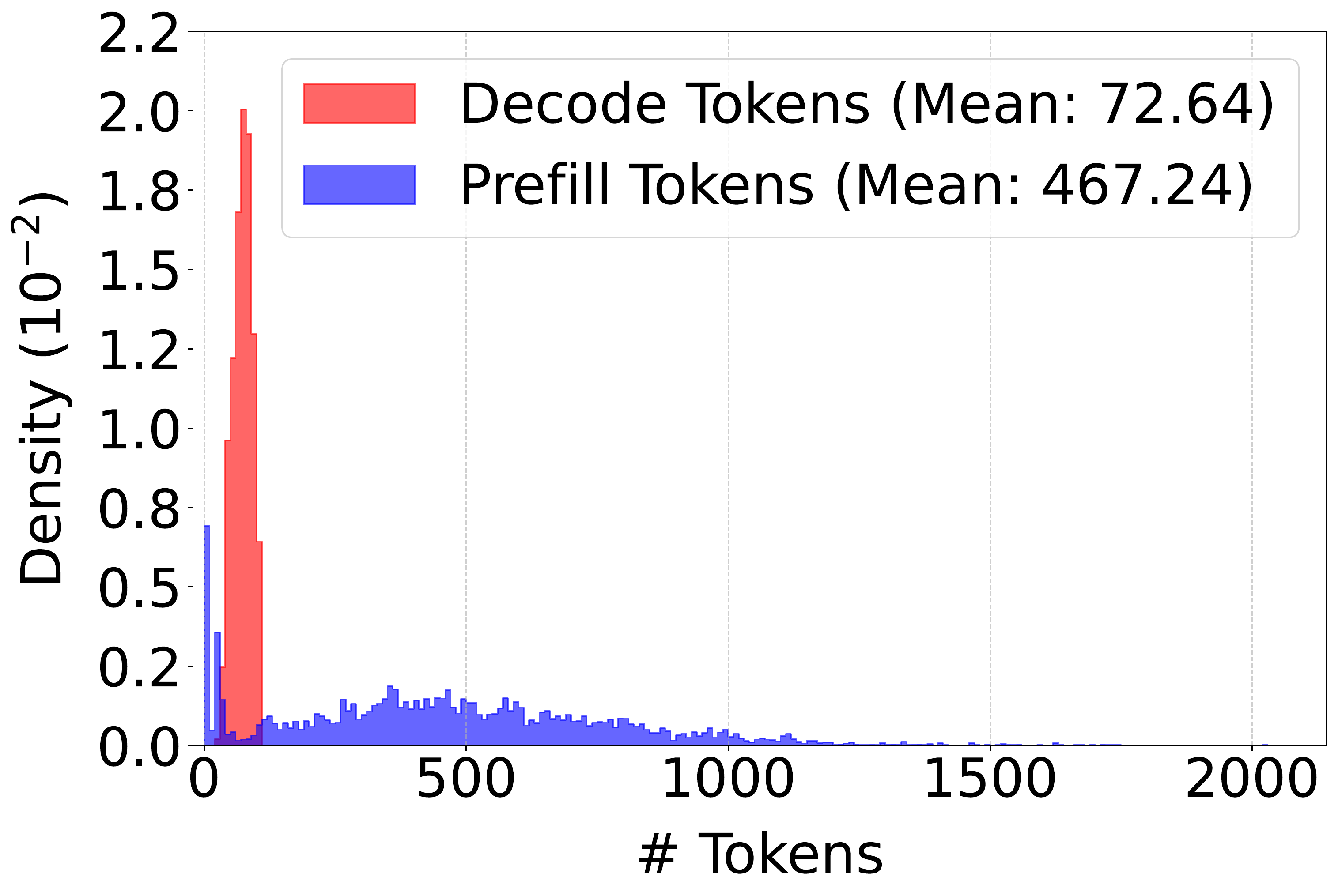}
    \caption{LATS}
    \label{fig:lats-llm-call-analysis}
\end{subfigure}
\hfill
\begin{subfigure}[b]{0.49\columnwidth}
    \includegraphics[width=\linewidth]{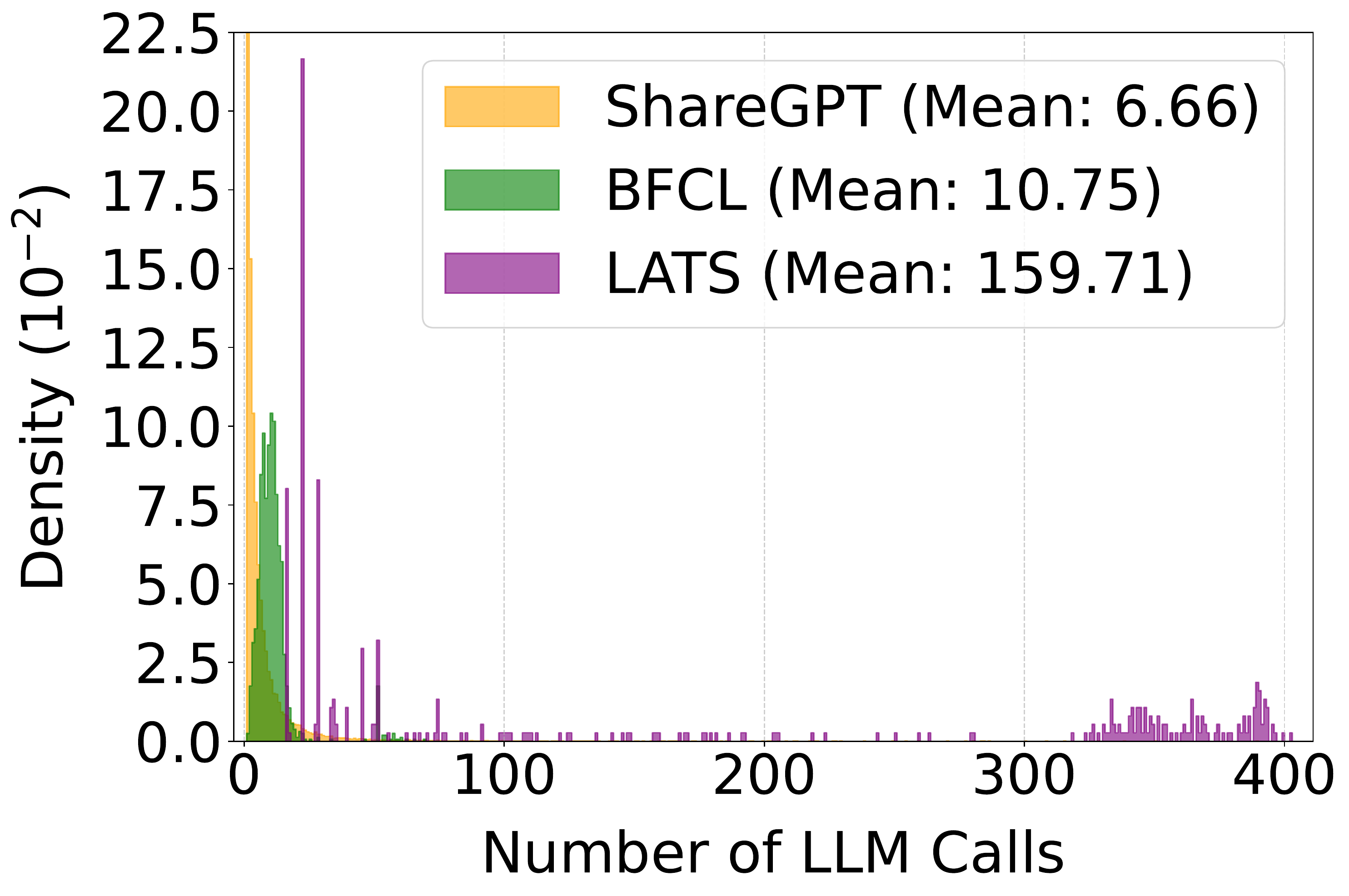}
    \caption{Number of LLM calls}
    \label{fig:num-llm-call-analysis}
\end{subfigure}
\caption{\small \textbf{Workload analysis.} LLM call statistics of programs from each workload. Input and output length distributions for (a) ShareGPT, (b) BFCL, and (c) LATS. Subfigure (d) plots the distribution of number of LLM calls in each workload.}
\label{fig:trace-analysis}
\vspace{-5mm}
\end{figure}
\begin{figure*}[!htb]
    \centering
\includegraphics[width=\textwidth]{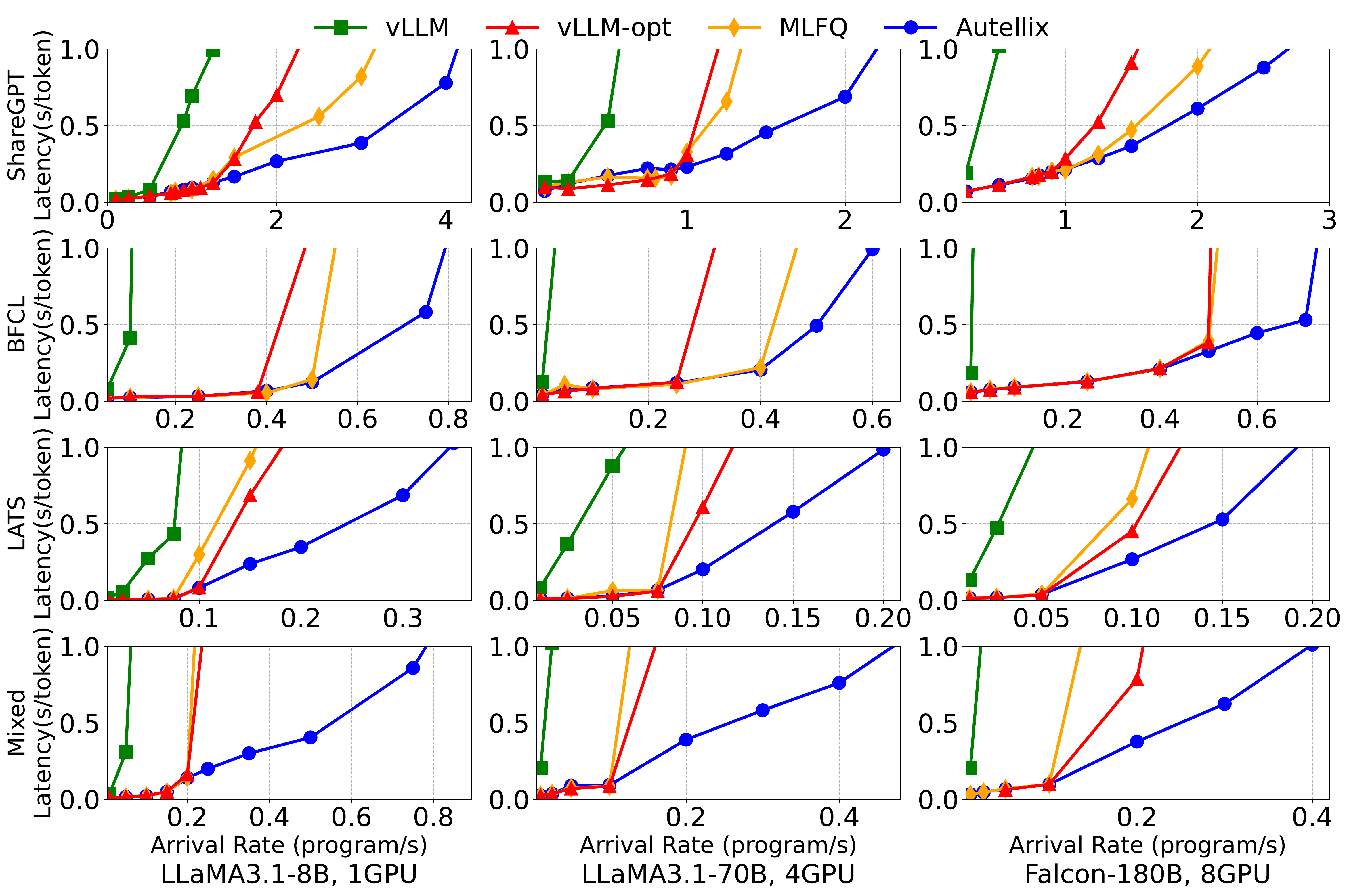} 
    \caption{\small \textbf{Single Engine, Main Results.} Average latency for different LLM serving systems across four real-world workloads.}  
    \label{fig:end2end_single_engine}
    \vspace{-4mm}
\end{figure*}

\begin{figure}[t]
    \centering
    \includegraphics[width=\columnwidth]{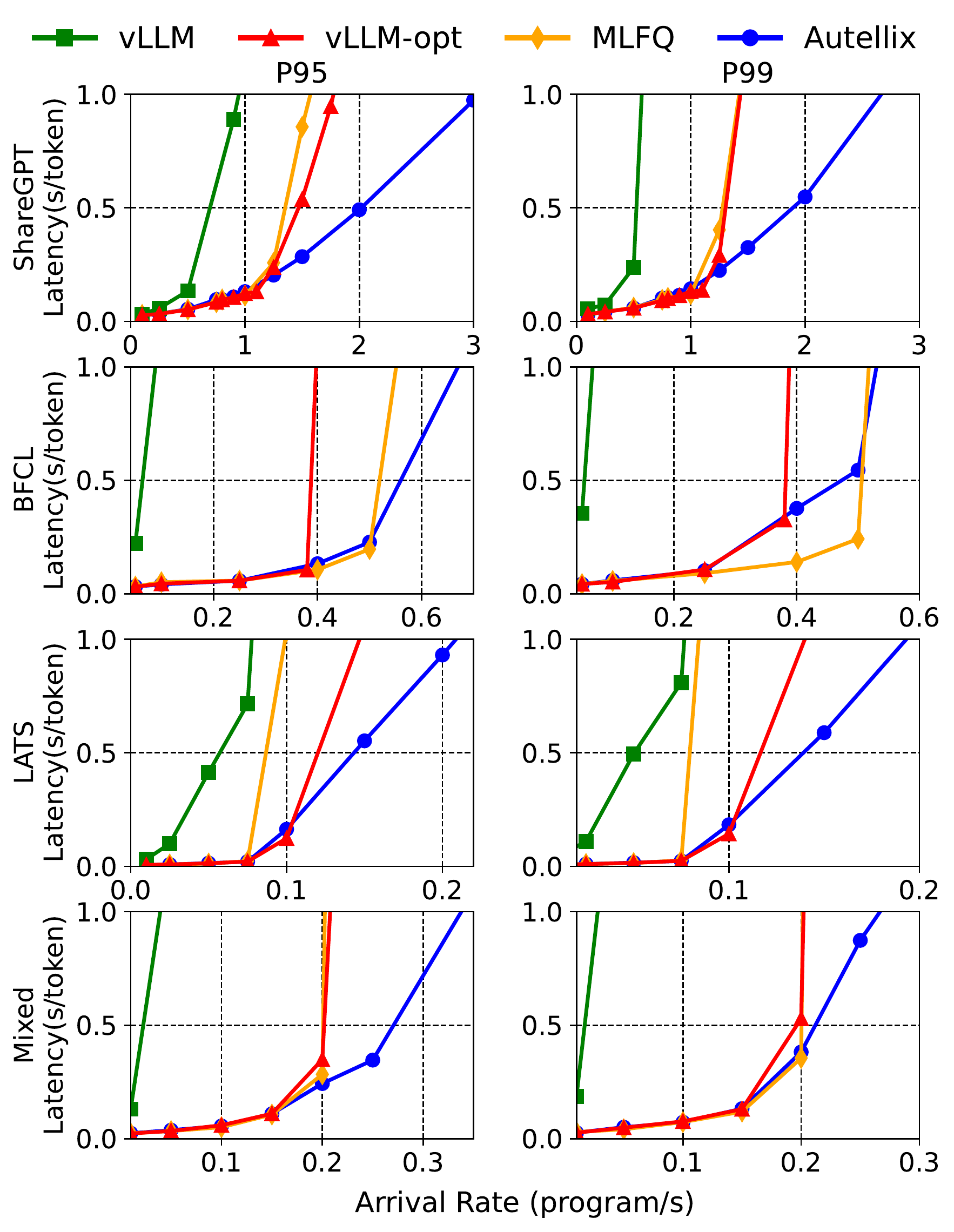} 
    \caption{\small \textbf{Single Engine, Tail Latencies.} 95\textsuperscript{th} (P95) and 99\textsuperscript{th} (P99) percentile latencies of different serving systems.}  
    \label{fig:singe-engine-95}  
    \vspace{-4mm}
\end{figure}

Our real-world experiments evaluate \text{\name} over four representative agentic workloads, which widely vary in the number of decode tokens, prefill tokens, and the LLM calls (Fig.~\ref{fig:trace-analysis}).

\vspace{1mm}
\noindent \textbf{Chatbot Agent: ShareGPT~\cite{sharegpt}.} The ShareGPT dataset comprises of user-generated conversational inputs and outputs, typical for chatbot applications. The number of LLM calls follows a long-tailed distribution with a mean of 6.66 and a max of 80 (Fig.~\ref{fig:num-llm-call-analysis}). ShareGPT's conversational nature is evident in its decode-heavy calls, averaging 277 decode tokens versus 256 prefill tokens, where short prompts generate detailed responses (Fig.~\ref{fig:sharegpt-llm-call-analysis}). Our experiments replay entire conversations as a program rather than the first turn.

\vspace{1mm}
\noindent \textbf{ReAct Agent: BFCL~\cite{berkeley-function-calling-leaderboard}.}
The Berkeley Function Calling Leaderboard (BFCLv3) evaluates LLMs on multi-turn, multi-step tool-usage tasks.  Compared to ShareGPT, BFCL's LLM calls are less long-tailed, with a mean of 10.75 and a maximum of 70 calls per program (Fig.~\ref{fig:num-llm-call-analysis}). BFCL is prefill-heavy, averaging 735.06 tokens per call due to long system prompts and detailed tool signatures, while decodes are short, averaging 34.14 tokens (Fig.~\ref{fig:bfcl-llm-call-analysis}). BFCL thus encapsulates dynamic workflows that alternate between heavy prefills phases and short decodes with function calls.

\vspace{1mm}
\noindent \textbf{Monte Carlo Tree Search: LATS~\cite{zhou2024languageagenttreesearch}.} LATS workloads, derived from running MCTS on HotpotQA~\cite{yang2018hotpotqadatasetdiverseexplainable}, are computationally intensive and involve many parallel LLM calls. Each program instance contains on average 159.7 LLM calls—an order of magnitude more than ShareGPT or BFCL workloads (Fig.\ref{fig:num-llm-call-analysis}). Moreover, the prefill and decoding phase of each call averages 467.2 and 72.6 tokens respectively (Fig.~\ref{fig:lats-llm-call-analysis}). These distributions highlight MCTS’s inherently iterative, parallel nature, pushing LLM serving systems to handle large volumes of concurrent calls efficiently. 

\vspace{1mm}
\noindent \textbf{Mixed.} We combine all three workloads, sampling equally from each to ensure diversity. This workload stress tests \text{\name}'s performance across different program classes.

\vspace{1mm}
\noindent For our experiments, we synthesize a trace by randomly sampling programs, not LLM calls, from the above workloads and generating programs' arrivals using a Poisson process $\lambda$, following established methodologies~\cite{vllm, fastserve}. This approach ensures our setup closely reflects real-world scenarios.

\subsection{Experimental Setup}

\noindent \textbf{Models \& Testbed.} We evaluate on three models: LLaMA-3.1-8B, 70B and Falcon-180B, running on 1, 4, and 8 GPUs, respectively. Experiments are conducted on a GCP Compute Engine \texttt{a2-ultragpu-8g} instance with eight A100-SXM4-80GB GPUs connected via NVLink, 1360 GB host memory, PCIe-4.0×16, and 2 TB of disk space.

\vspace{1mm}
\noindent \textbf{Metrics.} Existing LLM serving systems focus on request-level metrics, such as Time-to-First-Token (TFTT) and Time-per-Output-Token (TPOT), also referred to as token latency~\cite{vllm, fastserve, nanoflow}. However, these metrics overlook end-to-end latency for agentic programs. To that end, we introduce program-level token latency, defined as the total program response time divided by the number of tokens generated\footnote{For multi-threaded programs, \textit{program-level} token latency is computed as the critical path response time divided by the total tokens across all threads.}. A high-throughput system for programs should retain low program-level latency during high request rates. For simplicity, we refer to our metric as \textit{latency} throughout the evaluation.

\vspace{1mm}
\noindent \textbf{Baselines.} Our evaluation considers three baselines. All baselines, including \text{\name}, use the same max batch size.
\begin{itemize}[itemsep=0pt, parsep=0pt, topsep=0pt, partopsep=0pt, leftmargin=*]
    \item \textbf{vLLM~\cite{vllm}.} vLLM is the state-of-the-art, high throughput LLM serving system that integrates continuous batching~\cite{orca} and PagedAttention~\cite{vllm} to reduce KV cache fragmentation. Its default scheduling policy is FCFS, which is application-unaware and suffers from call-level and program-level HoL blocking. We use vLLM v0.6.1.
    \item \textbf{vLLM-opt.} An optimized version of vLLM that enables chunk-prefill~\cite{chunk_prefill}, prefix-caching~\cite{zheng2024sglangefficientexecutionstructured, lin2024parrotefficientservingllmbased}, and multi-step scheduling. Based on vLLM's blogpost~\cite{vllm2024perfupdate}, it's performance closely matches SGLang~\cite{zheng2024sglangefficientexecutionstructured} and TensorRT~\cite{NVIDIATensorRT-LLM}.
    \item \textbf{MLFQ.} On top of vLLM-opt, it implements preemption via the Multi-Level Feedback Queue algorithm~\cite{fastserve}. This baseline ablates the impact of program and call-level blocking.
\end{itemize}

\subsection{End-to-End Single-Engine Performance}
In Figure~\ref{fig:end2end_single_engine}, we evaluate the end-to-end performance of \text{\name} against three baselines and four workloads: ShareGPT, BFCL, LATS, and Mixed. Across all workloads, \text{\name} consistently achieves the highest throughput given same token latency. Conversely, vLLM performs worst due to its lack of prefix caching, which results in expensive recomputation of cumulative state (Fig.~\ref{fig:prefix_cache}) for LLM calls in the same program. Across workloads, the relative performance between vLLM-opt, MLFQ, and \text{\name} varies.

The first two rows plot the latencies for single-threaded workloads, ShareGPT and BFCL. vLLM and vLLM-opt’s FCFS scheduling causes severe head-of-line (HoL) blocking, which increases latencies as arrival rates increase. In contrast, MLFQ, a preemptive algorithm, mitigates call-level HoL, improving throughput by 1.5× over vLLM-opt. However, at high load, it still suffers from program-level HoL. By employing PLAS to tackle both call- and program-level HoL, \text{\name} achieves up to 8× throughput of vLLM, twice that of vLLM-opt, and a 1.5× improvement over MLFQ under heavy load.

The third row presents results for the multi-threaded LATS workload. \text{\name} outperforms vLLM, MLFQ, and vLLM-opt by up to 5×, 2.5×, and 2×, respectively. Notably, MLFQ’s preemptive scheduling, which benefits single-threaded programs, is less effective in multi-threaded settings. By aggressively prioritizing shorter requests, MLFQ inadvertently disrupts threads in the same program, exacerbating program-level HoL blocking and stalling overall progress. \text{\name}’s ATLAS policy holistically optimizes resource allocation across all threads, maintaining balanced progress and sustaining high throughput under heavy multi-threaded workloads.

The fourth row of Figure~\ref{fig:end2end_single_engine} illustrates performance on mixed workloads. \text{\name} achieves up to 15× higher throughput than vLLM, 5.5× higher than MLFQ, and 4× higher than vLLM-opt. Since \text{\name} reduces program and call-level blocking, \text{\name} performs better as programs' heterogeneity, or the diversity of LLM calls and decode lengths, increases. 




\begin{figure}[t]
    \centering
    \IfFileExists{plots/2engine1.pdf}{
        \includegraphics[width=\columnwidth]{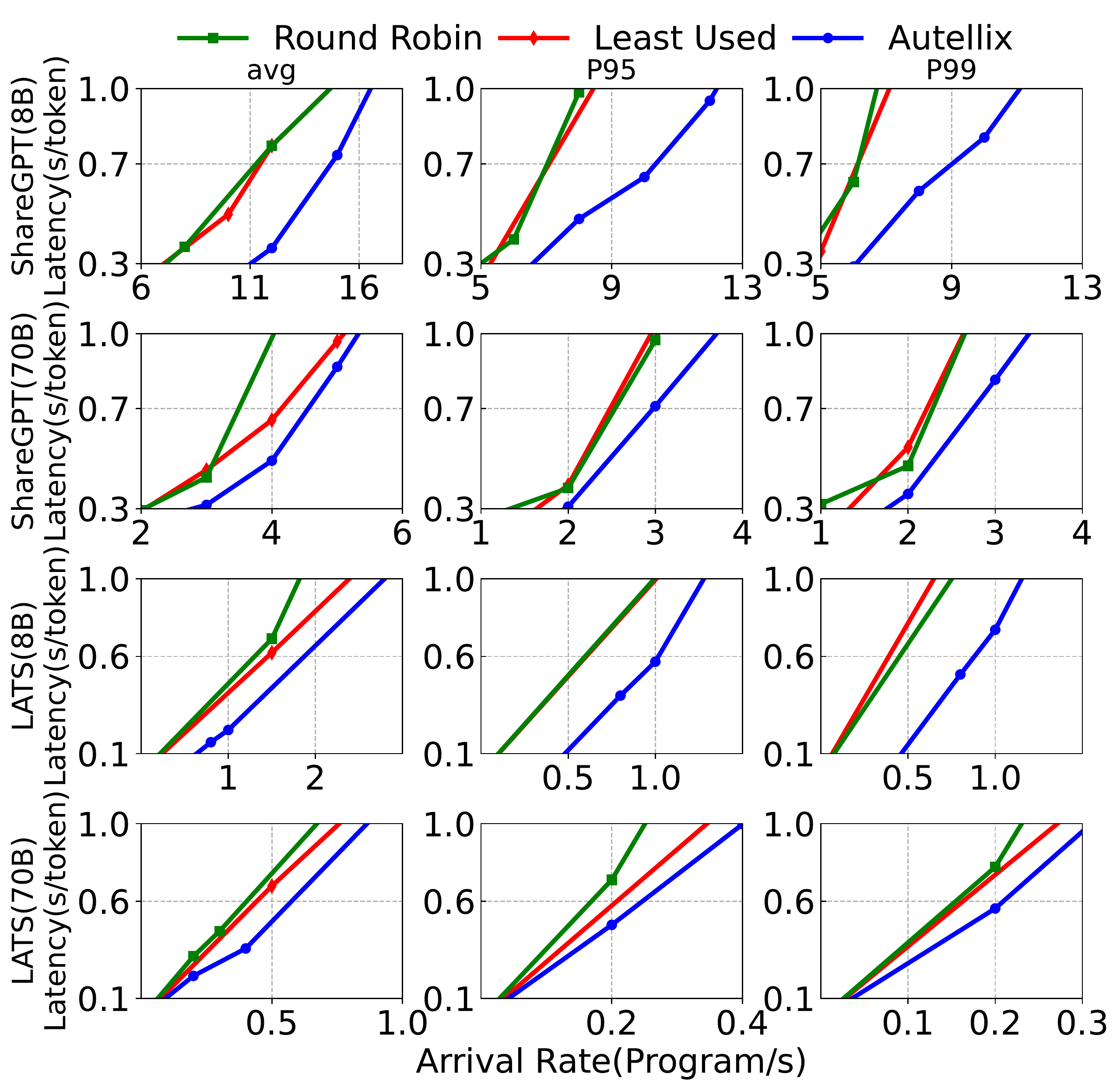}
    }{
        \fbox{
            \begin{minipage}[c][0.1\columnwidth][c]{0.95\columnwidth}
                \centering
                \textit{Placeholder for scalability.pdf}
            \end{minipage}
        }
    }
    \vspace{-4mm}
    \caption{\small \textbf{Multi-engine, Main Results.} Latencies (Avg., P95/99) w.r.t. different load balancing policies.} 
    \label{fig:two-engine} 
    \vspace{-2mm}
\end{figure}


\vspace{1mm}
\noindent \textbf{Tail latency.}
Preemptive scheduling strategies can reduce average latency but risk increasing tail latency by starving long-running programs. Figure~\ref{fig:singe-engine-95} reports the 95\textsuperscript{th} (P95) and 99\textsuperscript{th} (P99) percentile latencies across different workloads on LLaMA-3.1-8B. For ShareGPT, MLFQ significantly improves average latency compared to vLLM-opt (Fig.~\ref{fig:end2end_single_engine}), but exhibits poor P95/99 tail latencies. In contrast, for BFCL, MLFQ outperforms vLLM-opt in both cases. In 7 of 8 scenarios, \text{\name} maintains consistently lower tail latencies than MLFQ and vLLM-opt and improves throughput by up to 1.7× for P95/99 tail latencies, demonstrating robust performance gains in both average and tail performance metrics.

 

\subsection{End-to-End Multi-Engine Performance}   
To evaluate the effectiveness of \text{\name}'s data locality-aware load balancer (\S\ref{sec:load_balancer}), we compare it against two widely used load balancing strategies under identical scheduling policies (\textit{PLAS}, \textit{ATLAS}) for the sake of fairness:
\vspace{1mm}
\begin{itemize}[itemsep=0pt, parsep=0pt, topsep=0pt, partopsep=0pt, leftmargin=*] 
    \item \textbf{Round Robin.} Requests are assigned to engines in cyclic order—ensuring an even distribution of request counts—which is the default load-balancer policy for Kubernetes~\cite{kubernetes}. This strategy ignores data locality, resulting in costly KV cache misses and high recomputation overheads.
    \item \textbf{Least Used.} Requests are assigned to the engine with the lowest number of LLM calls in the system, effectively balancing engine workloads. However, like Round Robin, it neglects data locality and incurs frequent KV recomputations.
\end{itemize}

\vspace{1mm}

\noindent We conduct experiements using four replicas of LLaMA3.1-8B and two replicas of LLaMA3.1-70B with the ShareGPT and LATS workloads. The results, shown in Figure~\ref{fig:two-engine}, demonstrate the \text{\name}'s effectiveness in maintaining low average and tail latencies across all configurations. \text{\name} delivers up to 1.4× higher throughput compared to both baselines. The benefit is more pronounced in ShareGPT workload, where chat history reuse significantly amplifies KV-cache locality. These advantages become even more evident as the number of replicas increases, as a larger pool of engines reduces the likelihood of a request being routed to one with it's locality.

\begin{figure}[t]
    \centering
    \IfFileExists{plots/scalability.pdf}{
        \includegraphics[width=\columnwidth]{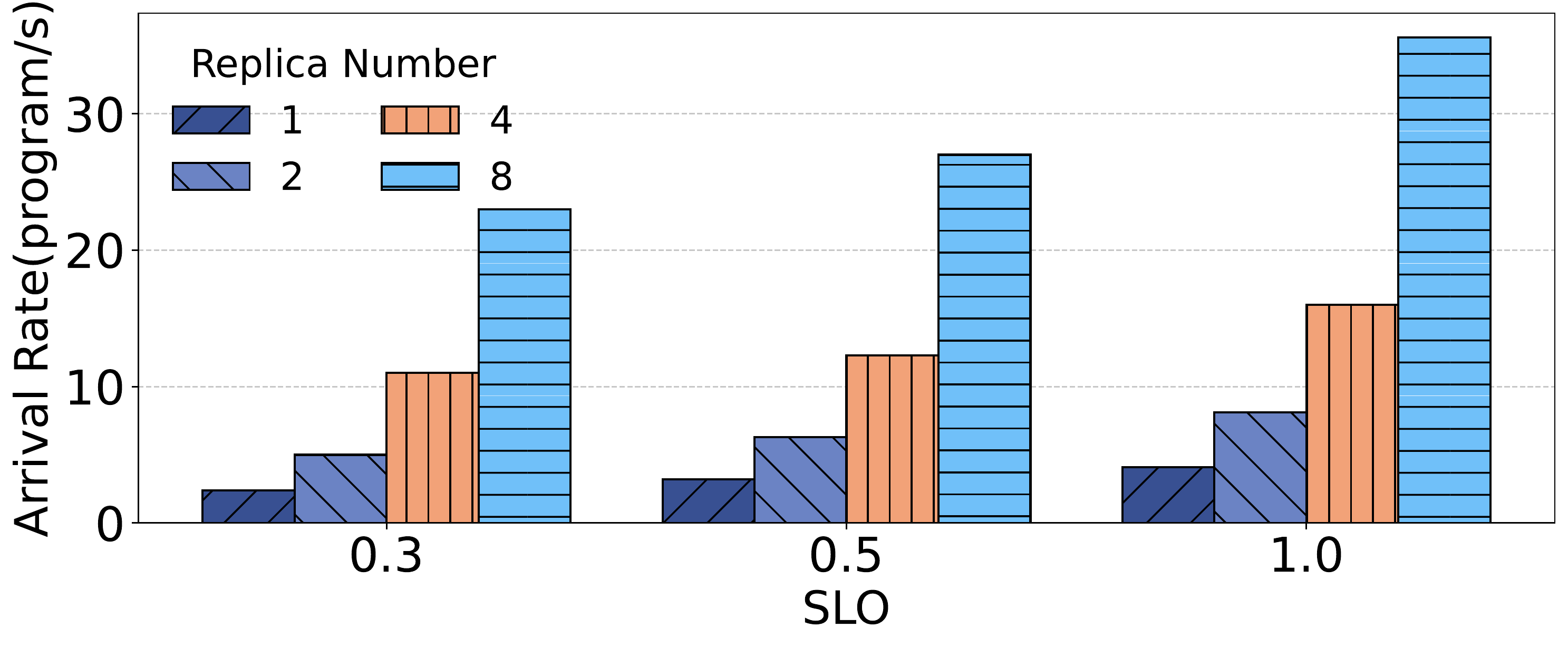}
    }{
        \fbox{
            \begin{minipage}[c][0.1\columnwidth][c]{0.95\columnwidth}
                \centering
                \textit{Placeholder for scalability.pdf}
            \end{minipage}
        }
    }
    \vspace{-4mm}
    \caption{\small \textbf{Scalability Experiments.} Given same SLO (defined as s/tok), \text{\name}'s max arrival rate (program/s) scales linearly w.r.t number of replicas, or LLM engines.}  
    \label{fig:scalability}  
\end{figure}

\vspace{1mm}
\noindent \textbf{Scalability.} To evaluate the scalability of \text{\name}, we assess its performance as the number of engine replicas increases under various latency requirements, using the ShareGPT workload with the LLaMA3.1-8B model. Figure~\ref{fig:scalability} shows linear scaling in all cases. Leveraging program-level load balancing, \text{\name} effectively scales horizontally without data locality overhead, making it a robust solution for large-scale LLM deployments.

\subsection{Ablations}

We ablate \text{\name} over different scenarios, including offine batch inference, timing breakdown, and various design choices, such as the swap kernel. All experiments run LLaMA3.1-8B~\cite{llama3} over ShareGPT~\cite{sharegpt} and LATS~\cite{zhou2024languageagenttreesearch}.

\vspace{1mm}
\subsubsection{Offline inference.}
\begin{figure}[t]
    \centering
    \IfFileExists{plots/makespan.pdf}{
        \includegraphics[width=\columnwidth]{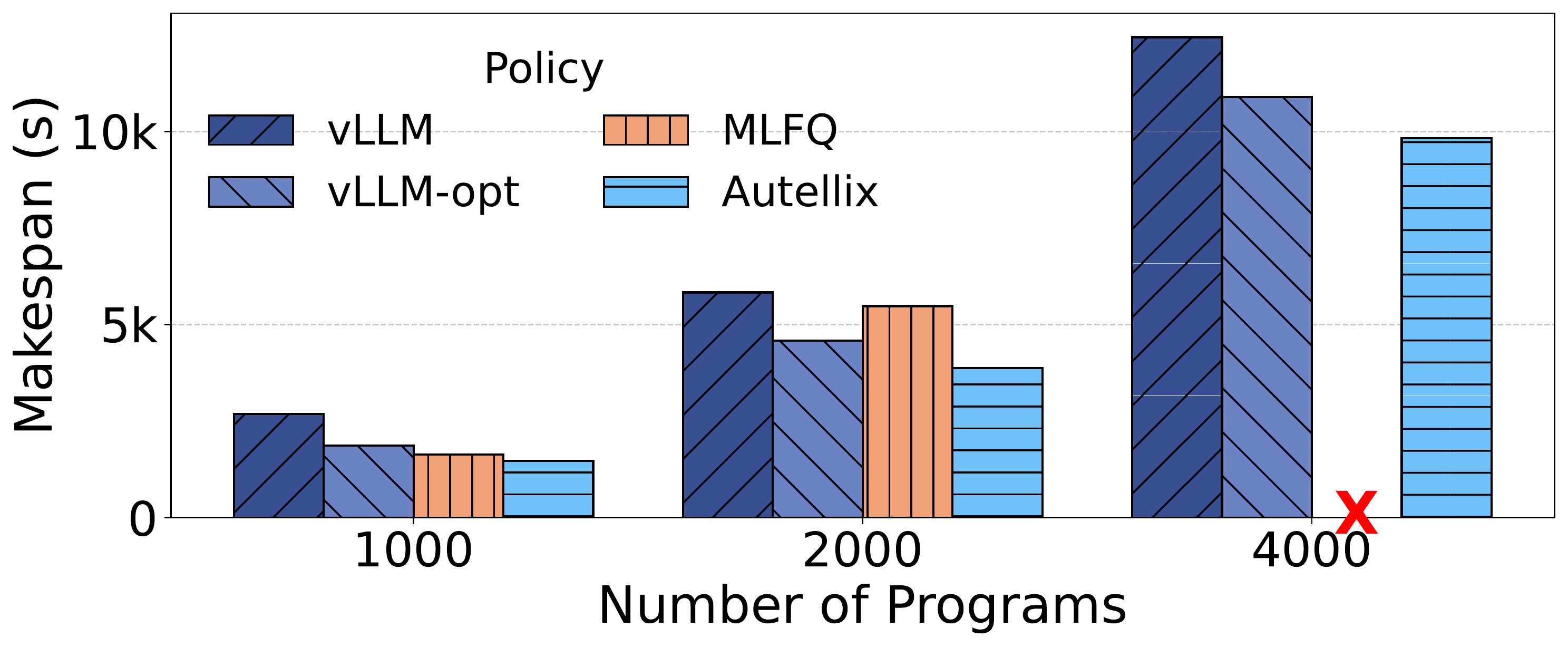}
    }{
        \fbox{
            \begin{minipage}[c][0.1\columnwidth][c]{0.95\columnwidth}
                \centering
                \textit{Placeholder for offline\_e2e.pdf}
            \end{minipage}
        }
    }
    \vspace{-6mm}
    \caption{\small \textbf{Offline batch inference.} \text{\name} decreases the time, or makespan, required to process a batch of programs.}  
    \label{fig:offline-single-engine} 
    \vspace{-2mm}
\end{figure}

In offline scenarios that prioritize throughput over latency, large batches of programs are processed in bulk rather than interactively or in a streaming fashion. We consider a use case where all programs are submitted at the start. Figure~\ref{fig:offline-single-engine} presents the makespan of all programs across all systems using the ShareGPT dataset. \text{\name} consistently outperforms the baselines, decreasing the average makespan by 10-40\%. At 4000 programs, MLFQ fails to complete execution. By assigning all new requests to the highest-priority queue, it creates many active LLM requests, causing severe memory contention and frequent GPU-CPU swapping. This overwhelms system resources, resulting in Out-Of-Memory (OOM) errors despite a large swap space (>1.2TB).

\subsubsection{Timing Breakdown}
\begin{figure}[t]
\centering
\begin{subfigure}[b]{0.49\columnwidth}
    \includegraphics[width=\linewidth]
    {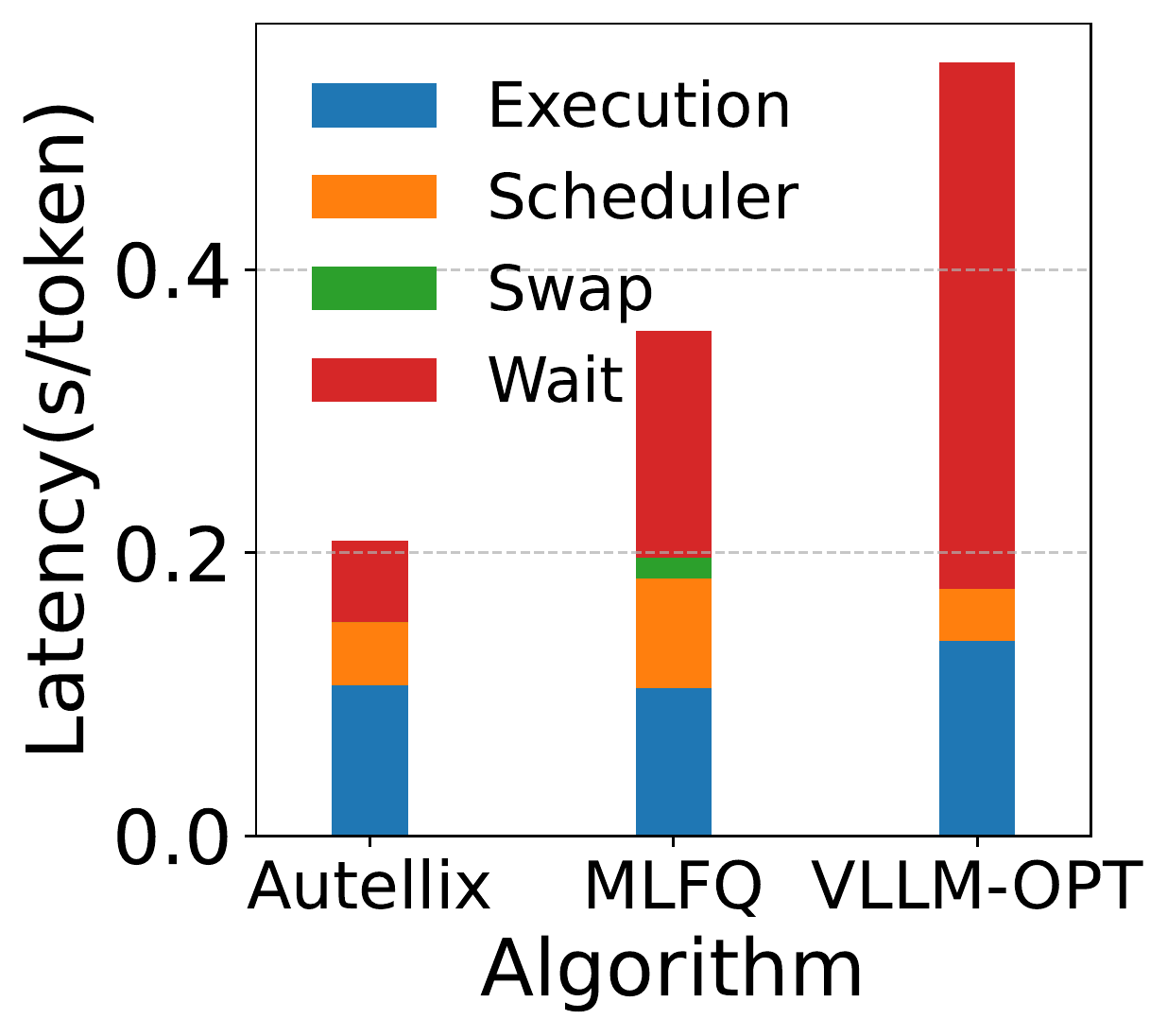}
    \caption{ShareGPT}
    \label{fig:Sharegpt-breakdown}
\end{subfigure}
\hfill
\begin{subfigure}[b]{0.49\columnwidth}
    \includegraphics[width=\linewidth]{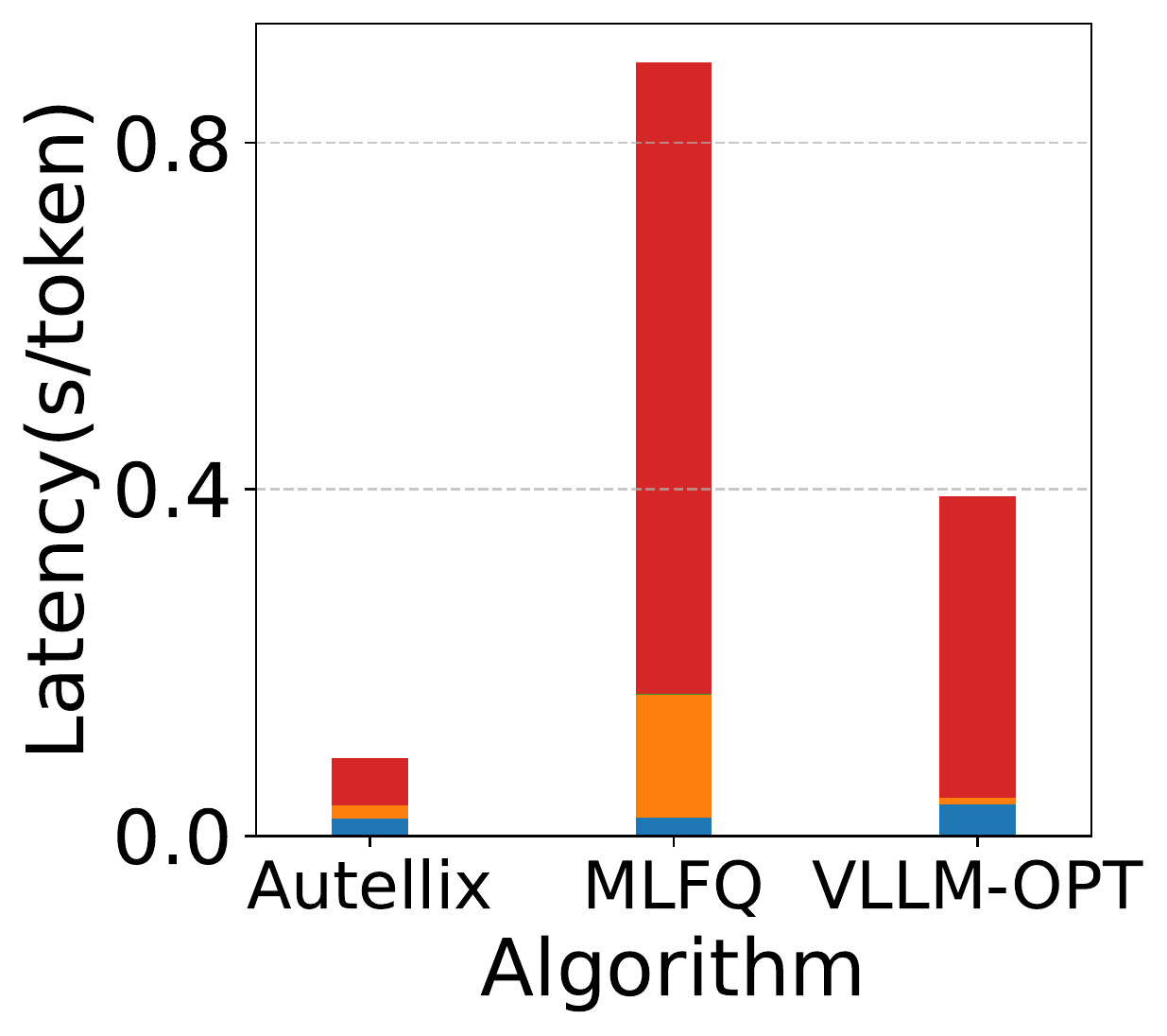}
    \caption{LATS}
    \label{fig:lats-break-down}
\end{subfigure}
\vspace{-2mm}
\caption{ \small \textbf{Breakdown of Inference Overheads.} \text{\name} significantly reduces wait time and introduces minor scheduler overheads to vLLM. \text{\name} also reduces swap times with its improved kernel.}
\label{fig:break_down}  
\vspace{-2mm}
\end{figure}
Figure~\ref{fig:break_down} breaks down the time LLM calls spend in the LLM serving layer for \text{\name} and its corresponding baselines. Overall, \text{\name} achieves lower token latency for ShareGPT and LATS by reducing wait and swap times, attributed respectively to \text{\name}'s program-level scheduling policy and improved swap kernels. Due to higher scheduling costs for preemption, both \text{\name} and MLFQ attain higher scheduling times than vLLM-OPT's naive FCFS. Yet, \text{\name} still incurs lower scheduling overhead than MLFQ by incorporating program-level priorities and better distributing LLM calls efficiently across different priority queues. In contrast, MLFQ assigns new LLM calls to the highest-level priority queue by default; hence, a majority of LLM calls reside in high-priority queues, leading to large scheduling overheads.

\subsubsection{Comparison to Optimal Scheduling}
\begin{figure}[t]
\centering
\begin{subfigure}[b]{0.49\columnwidth}
    \includegraphics[width=\linewidth]
    {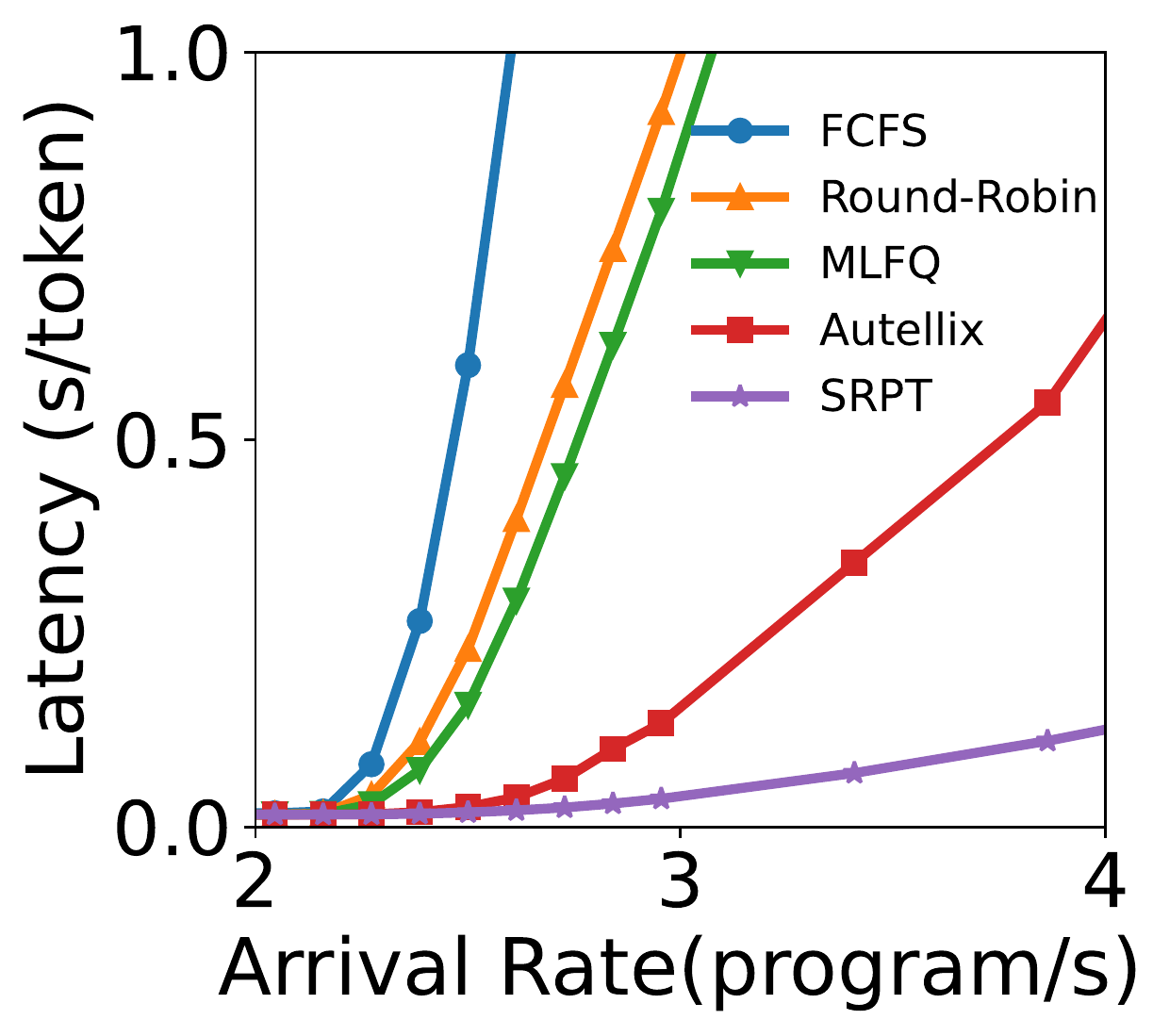}
    \caption{ShareGPT}
    \label{fig:sharegpt_sim}
\end{subfigure}
\hfill
\begin{subfigure}[b]{0.49\columnwidth}
    \includegraphics[width=\linewidth]{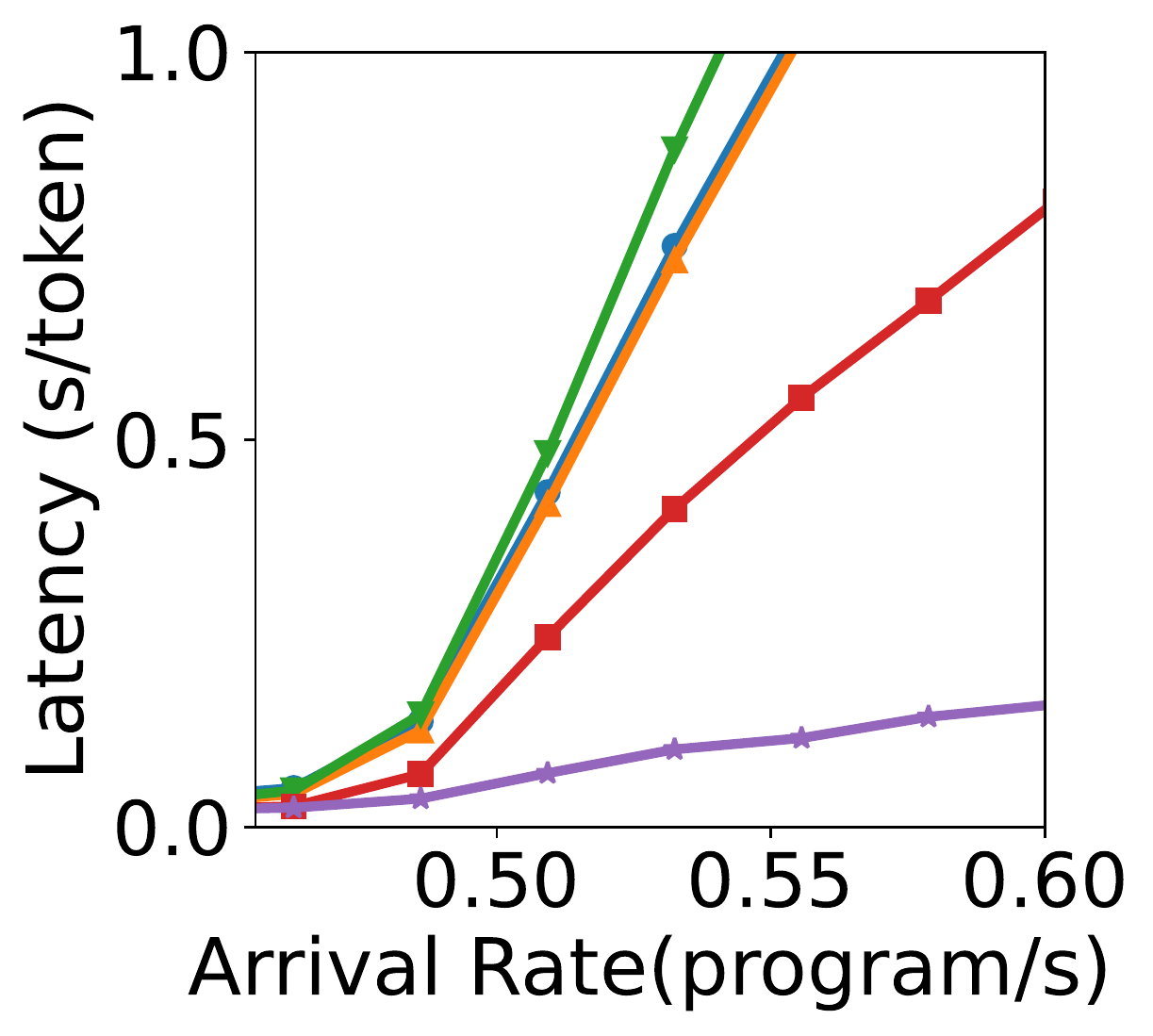}
    \caption{LATS}
    \label{fig:lats_sim}
\end{subfigure}
\vspace{-2mm}
\caption{ \small \textbf{Comparison to optimal scheduling policy.} In simulation, \text{\name} outperforms other scheduling policies; however, there remains a visible gap relative to the optimal policy (SRPT).}
\label{fig:sim}  
\vspace{-4mm}
\end{figure}
Optimal scheduling policies like Shortest Remaining Processing Time (SRPT) assume complete knowledge of each program’s runtime—an unrealistic assumption in practice. Hence, we emulate clairvoyance with a simulator by exposing each program’s total LLM calls and decode steps a priori. The simulation only considers scheduling, where each continuous-batching step is identical. Under these simplified conditions, \text{\name} outperforms FCFS and other preemptive schedulers (e.g., Round Robin, MLFQ). Nevertheless, a noticable gap remains between \text{\name} and SRPT, showing that prior knowledge can significantly boost performance.

\begin{figure}[t]
\centering
\begin{subfigure}[b]{0.49\columnwidth}
    \includegraphics[width=\linewidth]
    {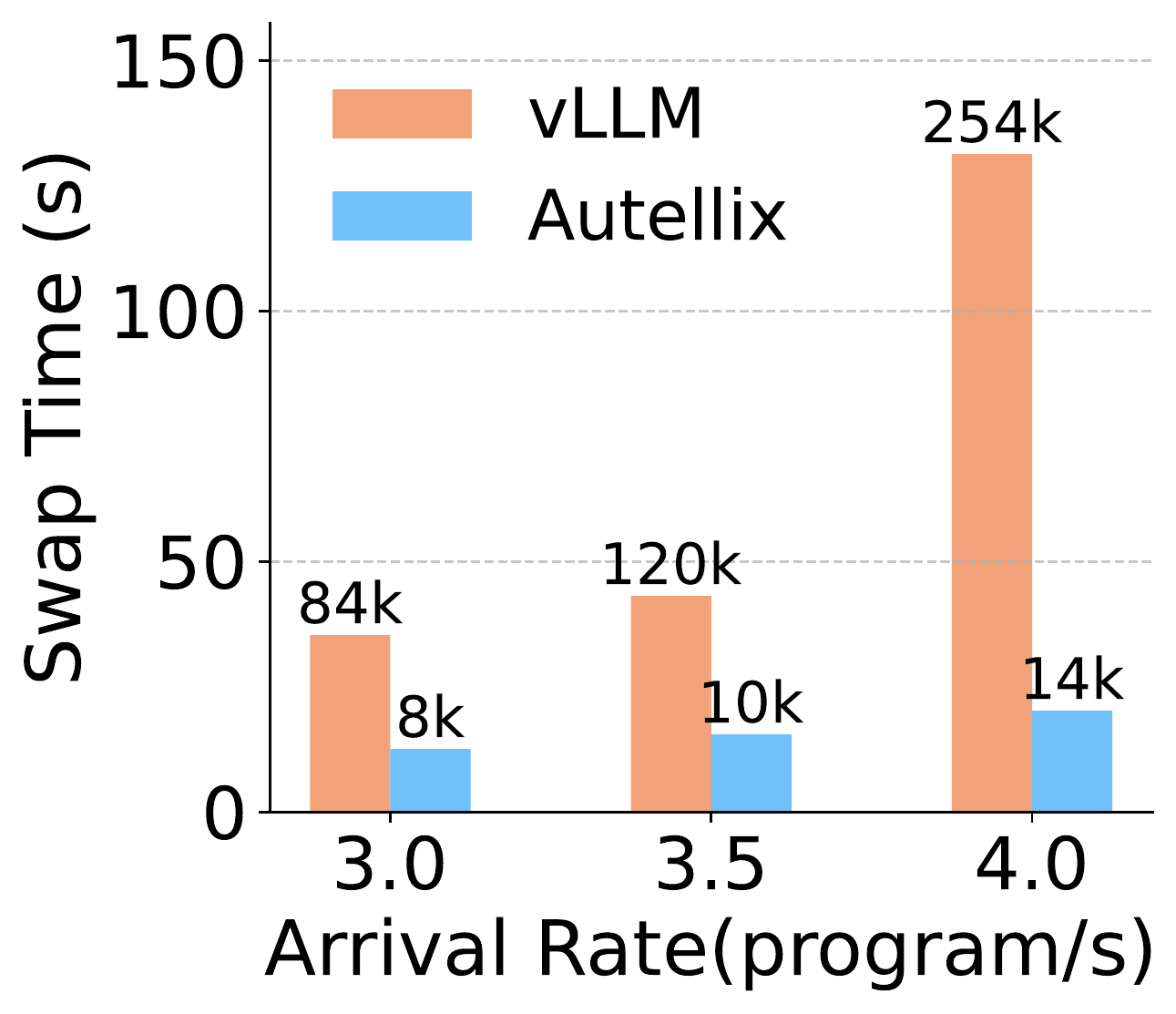}
    \caption{Swap Times}
    \label{fig:swap-kernel-time}
\end{subfigure}
\hfill
\begin{subfigure}[b]{0.49\columnwidth}
    \includegraphics[width=\linewidth]{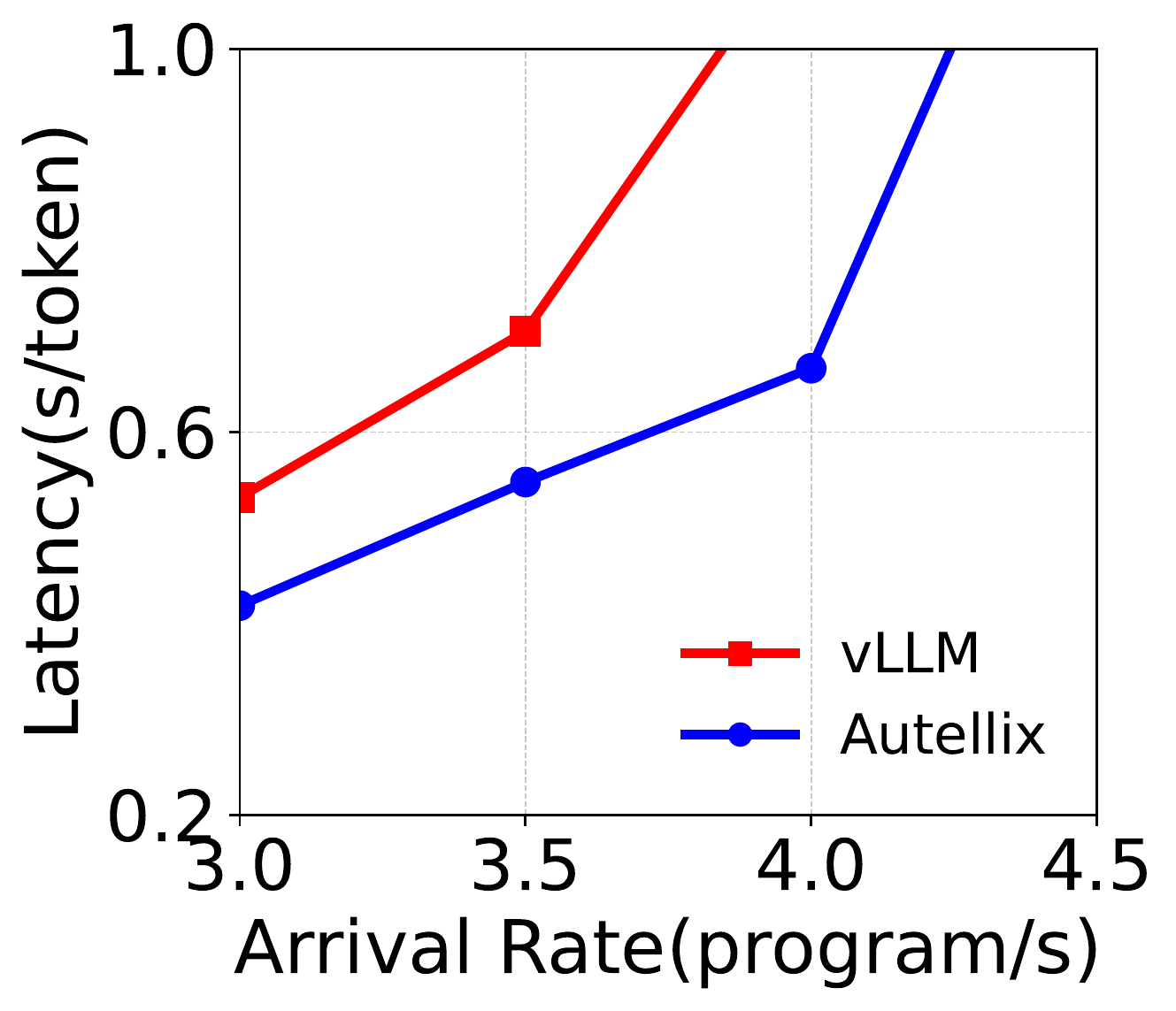}
    \caption{Throughput}
    \label{fig:swap-latency}
\end{subfigure}
\caption{ \small \textbf{Impact of \text{\name}'s swap kernel } \text{\name} reduces total swaps and GPU-CPU swap times, improving throughput.}
\label{fig:agentix_swap_sharegpt_8b}  
\vspace{-5mm}
\end{figure}

\vspace{1mm}
\subsubsection{Impact of Swapping Kernel} Preemptive scheduling increases active LLM calls in the system, incurring high GPU memory utilization. This leads to frequent GPU-CPU swaps for fetching relevant KV cache and significant swapping overheads at high request rates~\cite{fastserve}. \text{\name} mitigates this by batching parallel KV block transfers into a single operation—reducing swaps by up to 18x, swap times by 3-7x, and achieving 1.3x higher throughput than vLLM's implemented kernel (Fig.~\ref{fig:agentix_swap_sharegpt_8b}).

\section{Discussion \& Future Work}
\noindent \textbf{Graph Optimizations.} \text{\name} assumes no prior knowledge of a program’s execution DAG and dynamically constructs the graph as an internal representation (IR) during runtime. While full prior knowledge of a program’s execution is unrealistic, anticipating its immediate next steps can be practical—thereby enabling \textit{compiler optimizations} such as branch prediction and speculative execution, which enables future LLM calls to execute while prior calls are still completing. We defer such optimizations to future works.

\vspace{1.5mm}
\noindent \textbf{Post-Training.} Reasoning models, such as Deepseek-R1~\cite{deepseekai2025deepseekr1incentivizingreasoningcapability} and OpenAI's o1/o3 models~\cite{openai2025o1}, are post-trained via end-to-end reinforcement learning (RL) to optimize the thought process. To accelerate training, distributed RL systems alternate between distributed on-policy sampling and training to collect trajectories and perform policy gradient updates~\cite{sheng2024hybridflow,rllib}. With more effective scheduling, \text{\name} reduces the total makespan for batch sampling for each RL iteration, which immediately benefits distributed post-training systems.

\section{Conclusion}
\label{sec:conclusion}
We present \text{\name}, a distributed LLM serving system designed for highly-dynamic and general programs, not individual LLM calls. \text{\name}'s key innovation is to leverage program-level statistics, such as the cumulative service times, to better prioritize and schedule LLM calls, thereby improving the end-to-end response times and throughput of programs. We propose two general scheduling algorithms—for single- and multi-threaded programs—and a locality-aware load balancer that effectively reduces programs' waiting and execution times. Our experiments demonstrate that \text{\name} improves throughput of programs by 4×–15× at the same latency compared to state-of-the-art systems like vLLM.

\subsection*{Acknowledgement} We thank Pravein Kannan, Diana Arroyo, and Marquita Ellis from IBM for their insightful discussion. We thank Google Deepmind for funding this project, providing AI infrastructure for us to run experiments. Sky Computing Lab is supported by gifts from Accenture, AMD, Anyscale, Google, IBM, Intel, Microsoft, Mohamed Bin Zayed University of Artificial Intelligence, Samsung SDS, SAP, Uber, and VMware. 





\bibliographystyle{plain}
\bibliography{references}
\clearpage
\appendix

\clearpage
\end{document}